\documentclass{article} 
\usepackage{iclr2026_conference, times}

\usepackage{amsmath,amsfonts,bm}

\def\eqref#1{equation~\ref{#1}}

\def\1{\bm{1}}

\DeclareMathAlphabet{\mathsfit}{\encodingdefault}{\sfdefault}{m}{sl}
\SetMathAlphabet{\mathsfit}{bold}{\encodingdefault}{\sfdefault}{bx}{n}

\usepackage[utf8]{inputenc} 
\usepackage[T1]{fontenc}    
\usepackage{hyperref}       
\usepackage{url}            
\usepackage{booktabs}       
\usepackage{amsfonts}       
\usepackage{nicefrac}       
\usepackage{microtype}      
\usepackage{xcolor}  
\usepackage{amsmath}
\usepackage{amssymb}
\usepackage{mathtools}
\usepackage{amsthm}
\usepackage{enumitem}
\usepackage{multirow}
\usepackage{wrapfig}
\usepackage{subcaption}
\usepackage{tcolorbox}
\usepackage{array}
\usepackage{tabularx}
\usepackage{amsmath}
\usepackage{dsfont}
\usepackage[ruled,vlined,linesnumbered]{algorithm2e}
\usepackage[capitalize,noabbrev]{cleveref}

\newif\ifshowcomments
\showcommentstrue 

\ifshowcomments

\newcommand{\sw}[1]{{\footnotesize{\textcolor{orange}{[SW: {#1}]}}}\xspace}
\newcommand{\jzl}[1]{{\footnotesize{\textcolor{purple}{[JZL: {#1}]}}}\xspace}

\else

\newcommand{\sw}[1]{}
\newcommand{\jzl}[1]{}
\fi

\iclrfinaltrue

\title{EAMET: Robust Massive Model Editing \\
via Embedding Alignment Optimization}

\author{%
  Yanbo Dai, Zhenlan Ji, Zongjie Li$^{*}$, Shuai Wang\thanks{Corresponding authors.} \\
  Department of Computer Science and Engineering\\
  The Hong Kong University of Science and Technology \\
  \texttt{\{ydai851, zjiae, zligo, shuaiw\}@cse.ust.hk} \\
}

\begin{document}

\maketitle

\begin{abstract}
Model editing techniques are essential for efficiently updating knowledge in
large language models (LLMs). However, the effectiveness of existing approaches
degrades in massive editing scenarios, particularly when evaluated with
practical metrics. Their robustness is also limited in context-rich settings or
when editing multiple facts of the same subject simultaneously. We attribute
these failures to the embedding misalignment among knowledge items, which
undermines editing reliability at scale. To address this, we propose EAMET
(Embedding Alignment Model Editing in Transformers), which addresses this issue
by aligning the space of key and residual embeddings. Extensive experiments
across six LLMs and three datasets demonstrate that EAMET consistently
outperforms existing methods, achieving about 90\% editing efficacy when editing
10k facts. Codes and datasets are publicly available at
\url{https://github.com/ybdai7/EAMET-massive-editing}.
\end{abstract}

\section{Introduction}
\label{intro}
Large language models (LLMs) are increasingly employed as search engines and
chatbots, as they excel at retrieving knowledge to answer user
queries~\citep{brown2020language, touvron2023llama, yang2024qwen2,
bi2024deepseek}. However, they are prone to spreading misinformation about
frequently updated topics due to outdated training
data~\citep{vykopal2023disinformation, huang2025survey, xu2024hallucination}. To
address this issue, retraining or fine-tuning models for partial knowledge
updates is proposed~\citep{achiam2023gpt, team2025gemma}, albeit with
prohibitively expensive overhead. In contrast, recent advances in
locate-then-edit model editing (ME) techniques~\citep{meng2023memit,
fang2024alphaedit} enable massive editing of thousands of factual associations
concurrently at minimal data and computational cost, thereby rendering real-time
knowledge updates feasible.
\begin{figure}[!h]
    \begin{center}
    \vspace{-0.1in}
    \scalebox{0.9}{
    \centerline{\includegraphics[width=\columnwidth]{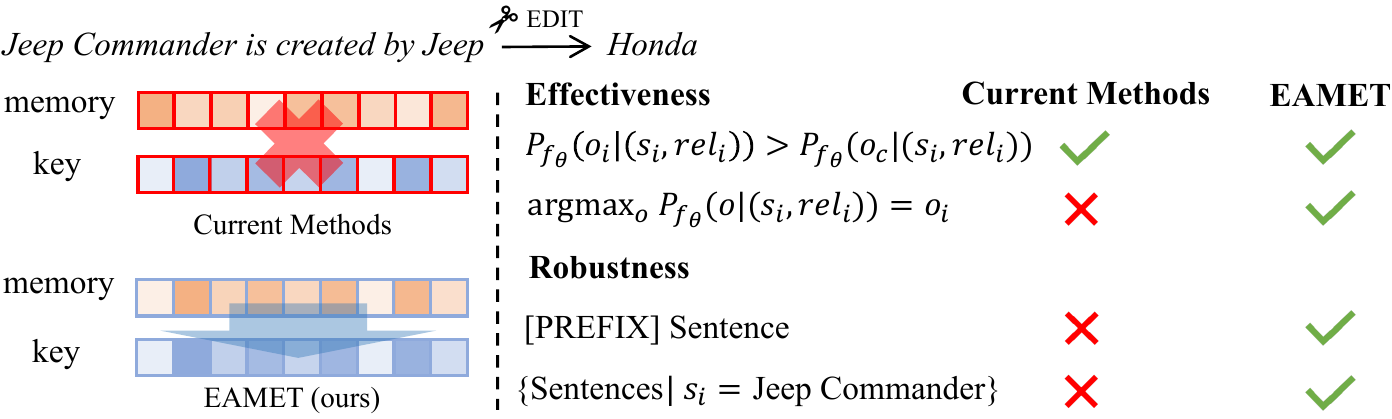}}
    } \caption{Illustration of current methods and our proposed EAMET in
    evaluating massive editing. Here, ``[PREFIX] Sentence'' and
    ``\{Sentence $\mid$ $s_i=\text{Jeep Commander}$\}'' denote the scenarios where
    the edited knowledge is preceded by prefixes and where multiple facts share
    the same subject, respectively.}
    \label{intro_fig}
    \end{center}
    \vspace{-0.15in}
\end{figure}

Despite the success of existing massive ME techniques, we observe that their
\emph{effectiveness} is often overestimated due to overly loose evaluation
metrics. In particular, most prior works assess editing quality by checking
whether the model is \textit{more likely} to generate the following tokens as
the target object than the original one, whereas neglecting to evaluate whether
the model's output is \textit{consistent with the target
object}~\citep{meng2023memit, fang2024alphaedit}. Therefore, we advocate a
``practical metric'', which measures the proportion of cases in which the edited
model retrieves the target object and explicitly generates related output. This
metric provides a more accurate reflection of real-world usage, as will be shown
in our evaluation setting (see \cref{exps}). Under such evaluation criteria,
existing methods fail to maintain their performance.

Moreover, existing methods exhibit limited \emph{robustness} in realistic
settings. We highlight two representative scenarios: (i) their performance
substantially degrades when edited knowledge is preceded by
prefixes~\citep{li2024unveiling}, a common phenomenon in practical
question-answering tasks~\citep{NEURIPS2024_d74033a2, NEURIPS2024_1568882b}; and
(ii) they fail to preserve accuracy when editing multiple facts associated with
the same subject, where performance drops markedly. Such lack of robustness in
massive editing scenarios undermines their applicability to real-world use
cases.

To analyze the limitations of existing methods, we first identify
\textit{``embedding misalignment''}, which reflects the structural inconsistency
between key and residual embedding spaces, as a primary factor underlying the
decline in both effectiveness and robustness during massive editing. Such
misalignment leads to information loss for individual knowledge updates. In
particular, when parameters are updated jointly from a batch of edited knowledge
items, they fail to accurately reconstruct an individual factual association.
This information loss becomes more severe as the number of edited items
increases.

To achieve effective and robust massive editing under practical settings, we
thus propose \textbf{EAMET} (\textbf{E}mbedding \textbf{A}lignment
\textbf{M}odel \textbf{E}diting in \textbf{T}ransformers), which outperforms
existing approaches under stricter evaluation criteria and exhibits strong
robustness in two described scenarios. EAMET addresses embedding misalignment by
progressively preserving optimized residual embeddings and aligning them with
the key embedding space, ensuring consistency throughout the editing process.

In this paper, we conduct extensive experiments on six LLMs, showing that EAMET
consistently surpasses existing methods under rigorous settings across the
CounterFact, ZsRE, and Wiki-recent datasets. EAMET maintains about 90\% editing
efficacy across all evaluated models and outperforms baselines by an average of
14\% and 8\%, with gains of up to 37\% and 15\% on CounterFact and ZsRE when
editing 10k facts. Moreover, EAMET sustains high accuracy even when edited
items are preceded by prefixes of up to 200 tokens or involve multiple facts
associated with the same subject. This demonstrates EAMET's robustness in
realistic and context-rich settings, including chatbots and long-context QA
tasks.

\section{Related Work}
\label{related_works}
\noindent \textbf{Model Editing.} Existing ME techniques can be classified into
auxiliary-based
\citep{NEURIPS2023_95b6e2ff,mitchell2022memory,zheng-etal-2023-edit, yu2024melo,
mitchell2022fast} and location-based methods \citep{meng2022locating,
meng2023memit, 10.1609/aaai.v38i17.29818}. Auxiliary-based ME techniques
preserve the original parameters, and introduce additional information to edit
knowledge. SERAC \citep{mitchell2022memory} requires extra memory to store new
edits and learn to reason over them to manipulate the model's output. Location-based
methods directly modify model parameters to edit knowledge without requiring any
additional information. These methods assume that factual associations are
stored in the feed-forward networks (FFNs) of the LLMs
\citep{geva-etal-2021-transformer, geva-etal-2022-transformer,
dai-etal-2022-knowledge}. Building on these, ROME \citep{meng2022locating} first
gains insights on the specific location of the knowledge through causal
analysis. It proceeds to directly modify critical MLP layers to update factual
associations. MEMIT \citep{meng2023memit} builds upon ROME to enable massive
editing of thousands of facts concurrently. AlphaEdit \citep{fang2024alphaedit}
focuses on sequential editing, aiming to preserve both previously edited
knowledge and the general capabilities of the LLM during successive edits.

\noindent \textbf{Massive Editing.} In practical applications, ME techniques may
aim to update a model with hundreds or even thousands of facts simultaneously in
order to keep up with the constantly evolving
knowledge~\citep{ju-etal-2024-investigating, gu-etal-2024-pokemqa}. However,
auxiliary-based methods are usually limited in scalability, typically supporting
only a few edits at a time \citep{mitchell2022memory}. In contrast,
location-based methods are more scalable for massive editing.
MEMIT~\citep{meng2023memit} scales to edit 10{,}000 facts concurrently, and
PMET~\citep{10.1609/aaai.v38i17.29818} further improves performance by
incorporating attention layers when updating the parameters of the FFNs. Despite
their effectiveness and scalability, these methods have been shown to be fragile
when handling \textit{prefixes} or multiple facts with the \textit{same subject}
during evaluation, which is a common scenario in real-world
applications~\citep{li2024unveiling, yang-etal-2024-butterfly,
ma2024robustnesseditinglargelanguage}. Moreover, we observe that their
performance in massive editing is overestimated due to the loose metric. In this
work, we propose EAMET, which achieves superior performance in massive editing
under practical evaluation metrics, while also exhibiting greater robustness
against long prefixes and multiple facts with the same subject.

\section{Preliminary: Editing Memory in LLMs}
\label{bkgrd}
Previous works have shown that a pre-trained LLM has memorized many factual
associations~\citep{petroni2019language, jiang2020can, roberts2020much,
shin2020autoprompt}. These stored facts could be edited by modifying the MLP
layers within FFN modules, based on the assumption that knowledge is stored in
them in the form of key-value pairs
\citep{geva-etal-2021-transformer,geva-etal-2022-transformer}. 

\begin{wrapfigure}{r}{0.55\textwidth}
    \vspace{-0.2in}
    \begin{center}
    \scalebox{0.55}{
    \centerline{\includegraphics[width=\columnwidth]{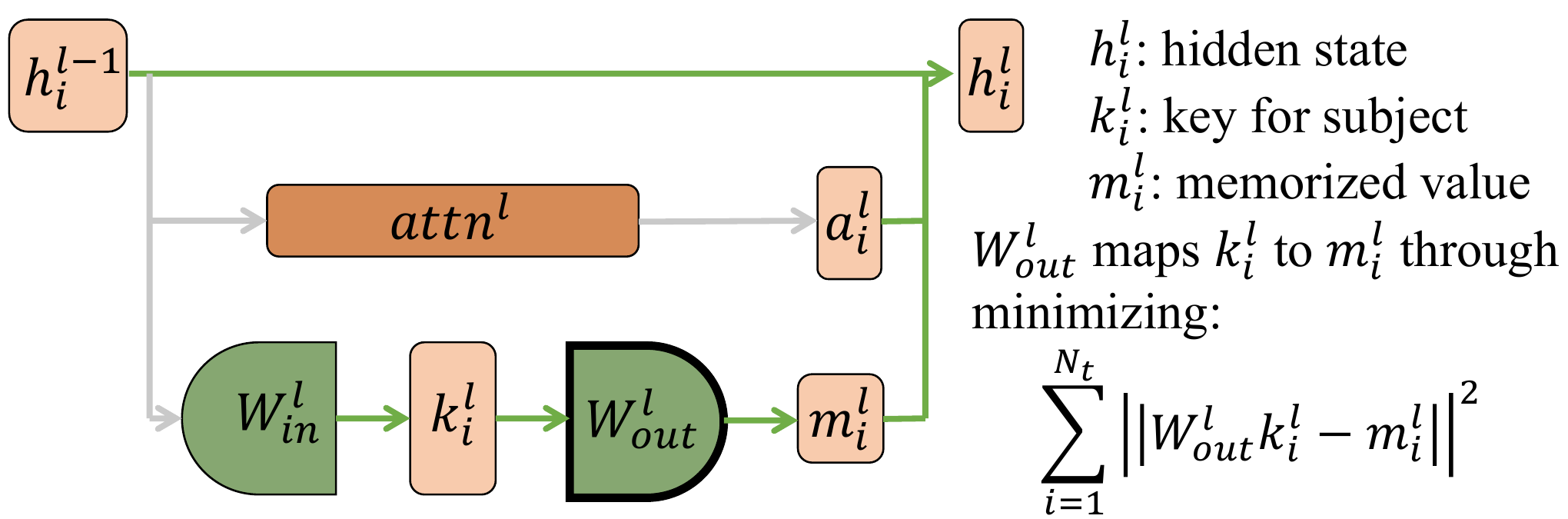}}
    }
    \caption{Illustration of the model editing problem.}
    \label{bkgrd_fig} 
    \end{center}
    \vspace{-0.2in}
\end{wrapfigure}

In \cref{bkgrd_fig}, the MLP layer $W_{out}^l$ within FFN associates
\textit{keys} $k_t^l(x)=\sigma(W^l_{in}\gamma(h^{l-1}_t(x)))$ with
\textit{memories} $m^l_t(x)$ for the fact $x$. Given the critical mediating role
of MLP layers in storing facts, Meng et al. \citep{meng2022locating}
shows that it is sufficient to update $W_{out}^l$ to edit stored facts.
We then optimize $W_{out}^l$ (abbreviated as $W_1$) as follows:
\begin{equation}
    \label{equation_optimze_final}
    W_1\triangleq\underset{\hat{W}}{\arg \min}(\sum_{i=1}^{N_t}||\hat{W}k_i^t-m_i^t||^2+\sum_{j=1}^{N_p}||\hat{W}k_j^p-m_j^p||^2)
\end{equation}
Here, $k_i^t$ and $k_j^p$ denote the encoded subject representations for
individual target and preserved fact $i$ and $j$, respectively, while $m_i^t$
and $m_j^p$ represent their corresponding memory vectors. We stack the keys and
memories of totally $N_t$ target knowledge into matrices as $K_t = [k_1^t\ |\
k_2^t\ |\ \dots\ |\ k_{N_t}^t]$ and $M_t = [m_1^t\ |\ m_2^t\ |\ \dots\ |\ m_{N_p}^t]$.
Similarly, we construct $K_p$ and $M_p$ for $N_p$ preserved facts. The objective
in \cref{equation_optimze_final} can then be optimized by solving the normal
equations~\citep{meng2023memit}:
\begin{align}
    \label{equation_block_form}
	(W_0+\Delta)\begin{bmatrix}K_p & K_t\end{bmatrix} &= \begin{bmatrix}M_p & M_t\end{bmatrix} \\
    \label{equation_w0}
	W_0K_p &= M_p
\end{align}
where we expand $W_1$ into $W_0 + \Delta$. $W_0$ denotes the original (unedited)
parameters that associate preserved keys with their memory representations. The
final update to $W_{out}^l$ can be computed by multiplying both sides of
\cref{equation_block_form} by $\begin{bmatrix}K_p & K_t\end{bmatrix}^T$, and
subtracting \cref{equation_w0} from
\cref{equation_block_form}\citep{meng2023memit}:
\begin{equation}
    \label{final_MLP_update}
    \Delta(C_p+K_tK_t^T)=RK_t^T
\end{equation}
where $R = M_t - W_0 K_t$ denotes new relations' residual with respect to the
original weights, which can also be written as $[r_1^t\ |\ r_2^t\ |\ \ldots\ |\
r_{N_t}^t]$.
Since the pretraining data of the original model is not accessible, we
approximate $C_p$ using a set of randomly sampled inputs from public datasets:
\begin{equation}
    C_p = \lambda E_{k^p}[k_i^p (k_i^p)^T]
\end{equation}
The scalar $\lambda$ balances the influence between newly edited facts and
preserved knowledge.

\section{Motivation}
\label{bkgrd_motivation}
In this section, we investigate the root causes of the challenges associated
with effective and robust massive editing, as illustrated in \cref{intro_fig}.
In particular, we analyze the decline in editing performance as the number of
edited facts increases. Our theoretical and empirical results indicate that
these issues arise from misalignment between key and residual embeddings. We
further examine robustness in two representative scenarios: (i) \emph{edits
preceded by long prefixes}, and (ii) \emph{edits applied to multiple facts
sharing the same subject}.

\subsection{Embedding Misalignment in Effective Massive Editing}
\label{motivation_theoretical}
\noindent\textbf{Theoretical Analysis.} We observe that by expanding $K_t$ and
$R$ in \cref{final_MLP_update}, the update equation can be reformulated as:
\begin{equation}
    \label{final_MLP_update_expanded}
    \Delta\left(C_p + \sum_{i=1}^{N_t} k_i k_i^T\right) = \sum_{i=1}^{N_t} r_i k_i^T
\end{equation}
where the update $\Delta$ is determined by the aggregated residual and key
embeddings across all edited facts. As the number of edits increases, solving
\cref{final_MLP_update_expanded} is more likely to cause reconstruction loss for
individual knowledge items due to the \emph{embedding misalignment} between the
residual and key embeddings. This eventually leads to degraded editing
performance.

To formalize the concept of embedding misalignment, we define two key
requirements for the desired update $\Delta$: (1) The update should preserve the
existing knowledge, expressed as $\Delta C_p = 0$. (2) The update should ensure
lossless reconstruction for each individual fact, formulated as $\Delta k_i=
r_i$, where $\Delta$ is computed while considering all target facts.
Incorporating (1), an ideal $\Delta$ that meets (2) implies:
\begin{equation}
    \label{collision_si}
    \Delta\left(C_p + \sum_{i=1}^{N_t} k_i k_i^T\right) = \sum_{i=1}^{N_t} r_i k_i^T \quad \rightarrow \quad \Delta k_i = r_i \quad \text{for } i = 1, 2, \ldots, N_t
\end{equation}
However, the validity of \cref{collision_si} is intuitively affected by the
degree of misalignment between the residual and key embedding of different facts.
We then define embedding misalignment:

\noindent \textbf{Definition 1 (Embedding Misalignment).} \textit{Given $N$
knowledge items, let each item $i$ be associated with a residual embedding $r_i$
and a key embedding $k_i$. We define the embedding misalignment of item $i$
as the structural similarity between the pairwise relations of its residual
embedding and those of its key embedding. Formally, consider the distributions
\begin{equation}
P_r^{(i)} = \{\, cos(r_i, r_j) \mid j \ne i \,\}, 
\qquad
P_k^{(i)} = \{\, cos(k_i, k_j) \mid j \ne i \,\},
\end{equation}
where $cos(\cdot,\cdot)$ is the cosine similarity. The $i$th misalignment 
score is quantified by the KL divergence:
\begin{equation}
    \label{embedding_misalignment}
    \mathcal{A}(i) = \mathrm{KL}\!\left( P_r^{(i)} \;\|\; P_k^{(i)} \right).
\end{equation}}

We now formalize the connection between embedding misalignment and the editing
performance of a specific knowledge item $i$ under massive editing.
Specifically, we quantify the degree to which \cref{collision_si} is established
by analyzing the reconstruction loss $e_i = \Delta k_i - r_i$ for each
knowledge item. This relationship is formalized in the following theorem:

\noindent\textbf{Theorem 1.}
Let $\Delta$ be the closed-form solution satisfying 
$\Delta \sum_i k_i k_i^\top = \sum_i r_i k_i^\top$, and define the reconstruction
residual of item $i$ as $e_i = \Delta k_i - r_i$. Then we can expand
\begin{equation}
e_i = \sum_{j=1}^N \beta_{ij} r_j - r_i,
\qquad 
\beta_{ij} := k_j^\top\!\Big(\sum_{\ell=1}^N k_\ell k_\ell^\top\Big)^{\!-1}\! k_i
\end{equation}
and its norm is bounded by the misalignment between the neighborhood structures of
$r_i$ and $k_i$:
\begin{equation}
\|e_i\|
\;\le\;
C_i \sqrt{\tfrac{1}{2}\,\mathcal{A}(i)}
\;+\; |\beta_{ii}|\,\|r_i\| \;+\; \|\varepsilon_i\|,
\end{equation}

This result demonstrates how embedding misalignment impacts the editing performance
of individual knowledge items under massive editing. Specifically, stronger
misalignment among knowledge items leads to increased individual reconstruction
loss, ultimately reducing the overall effectiveness of massive editing. The
complete proof is provided in \cref{app:proof_theorem1}.

\begin{wraptable}{!h}{0.6\textwidth}
    \caption{Editing performance with varying numbers of edited facts on
    LLaMA2-7B and Deepseek-7B.}
    \vspace{-0.1in}
    \scalebox{0.8}{
        \begin{tabular}{c||ccc|ccc}
        \toprule
        \multicolumn{1}{c}{\textbf{Model}} & \multicolumn{3}{c}{\textbf{LLaMA2-7B}} & \multicolumn{3}{c}{\textbf{Deepseek-7B}}\\
        \cmidrule(lr){2-4}\cmidrule(lr){5-7}
        \multicolumn{1}{c}{\textbf{No. of Edited Facts}} & \textbf{200} & \textbf{500} & \textbf{1000} & \textbf{200} & \textbf{500} & \textbf{1000} \\
        \midrule
        Editing Efficiency(\%) & 98.5 & 90.0 & 86.8 & 99.5 & 98.6 & 97.8 \\
        $\sum_i \mathcal{A}(i)$ & 79 & 243 & 554 & 68 & 223 & 562 \\
        \bottomrule
        \end{tabular}
    }
    \label{tab:editing_stats}
\end{wraptable}

\noindent\textbf{Empirical Study.} Motivated by the above analysis, we
hypothesize that the failure of massive editing stems from misalignment between
the embeddings of different knowledge items. To test this hypothesis, we edit
200, 500, and 1,000 facts from the CounterFact dataset~\citep{meng2022locating}
using MEMIT~\citep{meng2023memit} on
LLaMA2-7B~\citep{touvron2023llama2openfoundation} and
Deepseek-7B~\citep{bi2024deepseek}. We then evaluate the editing accuracy of
these items when no prefix is added to the edited query. Embedding misalignment is
quantified using the misalignment score defined in \cref{embedding_misalignment}.

As shown in \cref{tab:editing_stats}, the overall editing accuracy of both
models decreases as more facts are edited, accompanied by a clear increase in
embedding misalignment. For example, on LLaMA2-7B, the accuracy drops from
98.5\% to 86.8\% as the number of edited facts grows from 200 to 1,000, while
the misalignment score rises from 79 to 554. These results provide further
evidence for our theorem that embedding misalignment leads to degraded editing
performance.

\subsection{Impact of Embedding Misalignment on Editing Robustness}
We investigate how embedding misalignment affects robustness in massive editing
along two dimensions: (i) long-prefix perturbations and (ii) simultaneous edits
of samples sharing the same subject. Based on our theoretical and empirical
analysis, we derive two corollaries to characterize these effects and validate
them with controlled experiments.

\textbf{Corollary 1.} \textit{Long prefixes exacerbate embedding misalignment
issues under massive editing, leading to degraded editing performance when
edited facts are evaluated with descriptive prefixes.}
\begin{table}[!h]
    \begin{center}
    \vspace{-0.1in}
    \caption{Impact of varying prefix lengths on editing performance.}
    \label{tab:concurrent_edits}
    \scalebox{0.8}{
    \begin{tabular}{c||cccccccc}
        \toprule
        \multicolumn{1}{c}{\textbf{Model}} & \multicolumn{4}{c}{\textbf{LLaMA2-7B}} & \multicolumn{4}{c}{\textbf{Deepseek-7B}} \\
        \cmidrule(lr){2-5}\cmidrule(lr){6-9}
        \multicolumn{1}{c}{\textbf{Prefix Lens}} & \textbf{0} & \textbf{5} & \textbf{10} & \textbf{50} & \textbf{0} & \textbf{5} & \textbf{10} & \textbf{50} \\
        \midrule
        Editing Acc. & 98.50\% & 84.15\% & 80.35\% & 77.40\% & 99.50\% & 90.35\% & 84.25\% & 85.10\% \\
        low $\mathcal{A}(i)$ Acc. & - & 94.00\% & 91.00\% & 90.00\% & - & 94.00\% & 92.00\% & 91.00\% \\
        top $\mathcal{A}(i)$ Acc. & - & 46.00\% & 46.00\% & 45.00\% & - & 55.00\% & 54.00\% & 47.00\% \\
        \bottomrule
    \end{tabular}
    }
    \end{center}
    \vspace{-0.2in}
\end{table}

\textbf{Empirical Verification of Corollary 1.} We edit 200 CounterFact facts on
LLaMA2-7B and Deepseek-7B using MEMIT, and evaluate average editing accuracy
with prefix lengths ranging from 5 to 50 tokens. To assess the impact of
embedding misalignment, we also compare the 10 items with the highest and lowest
misalignment scores under long-prefix conditions.

\cref{tab:concurrent_edits} shows that editing performance degrades when edited
facts are evaluated with prefixes. For LLaMA2-7B, accuracy falls from 98.5\% to
84.15\% with a 5-token prefix, and further to 77.40\% with a 50-token prefix. A
similar trend is observed for Deepseek-7B, where editing accuracy drops by
around 15\% under 50-token prefixes. The robustness to prefix perturbations
varies markedly between items prone to embedding misalignment and those that are
not. Consistent with \textbf{Corollary 1}, items with lower misalignment scores
maintain above-average accuracy, whereas highly misaligned items suffer a sharp
decline, with accuracy dropping to an average of 48\%.

\textbf{Corollary 2.} \textit{Massive editing suffers from degraded performance
when multiple samples with the same subject are edited simultaneously. In this
case, the reconstruction weight on the target $\beta_{ii}$ decreases while
cross-weights $\beta_{ij}$ increase, eventually leading to reconstruction
failure of $r_i$.}

\begin{figure}[!h]
    \begin{center}
    \vspace{-0.1in}
    \caption{Impact of editing same-subject samples. Shaded region indicates shared items.}
    \vspace{-0.1in}
    \label{fig:multiple_subjects}
    \includegraphics[width=0.9\columnwidth]{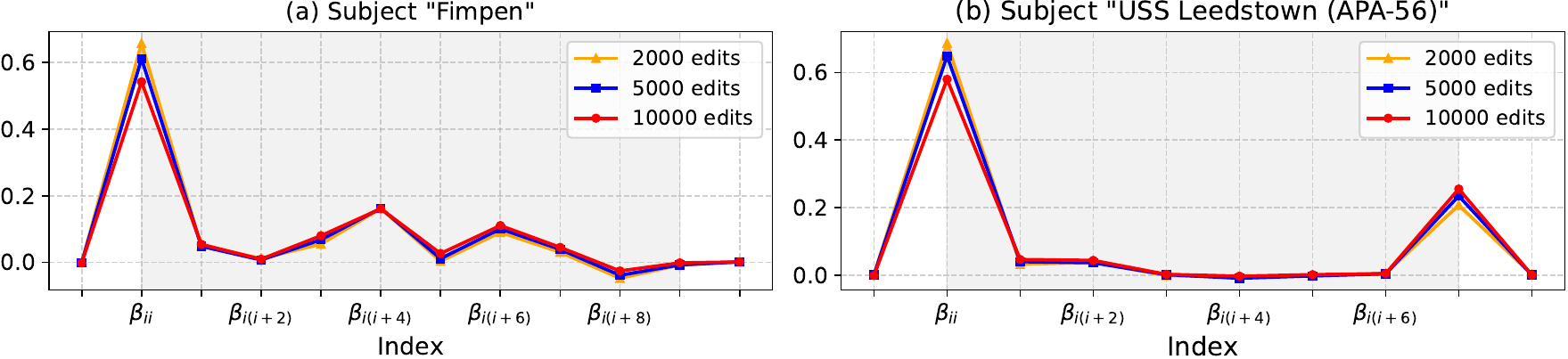}
    \vspace{-0.1in}
    \end{center}
\end{figure}

\textbf{Empirical Verification of Corollary 2.} 
Figure~\ref{fig:multiple_subjects} shows the reconstruction coefficients for two
example subjects under different numbers of edits. Although $\beta_{ii}$
remains the dominant coefficient, its value decreases steadily as the number of
co-edited samples increases, while off-diagonal coefficients $\beta_{ij}$ grow
accordingly. As a result, $\Delta k_i$ is no longer primarily aligned with $r_i$
but is instead reconstructed as a mixture of other $r_j$, making the recovery of
the correct target representation increasingly difficult.

This behavior is consistent with the misalignment measure $\mathcal A(i)$, as
only $\beta_{ij}$ from the same subject as $i$ take relatively large
values, while cross-subject weights remain negligible. Therefore, reconstruction
is dominated by the same-subject neighborhood. When $\mathcal A(i)$ is small,
same-subject embeddings are well aligned across both $k$-space and $r$-space.
Thus, using other $r_j$ from the same subject to approximate $r_i$ introduces
only limited error. However, when $\mathcal A(i)$ is large, misalignment within
this neighborhood amplifies the effect of weight redistribution, causing the
residual $\|e_i\|$ to grow and ultimately leading to degraded editing
performance. We provide details in \cref{app:same-subject-analysis}.

These findings underscore the strong connection between embedding misalignment
and the effectiveness as well as robustness of massive editing. Motivated by
this observation, the following section introduces our approach for aligning
key and residual embeddings to enhance the overall performance of massive
editing.

\section{Embedding Alignment Memory Optimization}
\label{methods}
\begin{figure}[!h]
    \vspace{-0.1in}
    \begin{center}
    \scalebox{1}{
    \centerline{\includegraphics[width=\columnwidth]{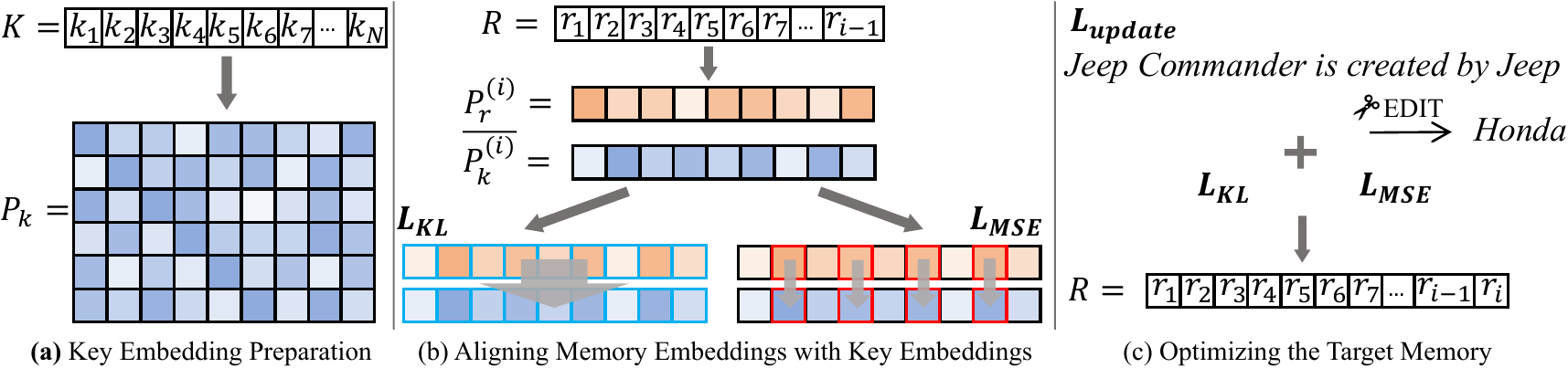}} }
    \caption{Method Overview of EAMET. }
    \label{method_overview}
    \end{center}
    \vspace{-0.1in}
\end{figure}
Motivated by these results, we propose \emph{EAMET}, which optimizes memory
embeddings to promote alignment with key embeddings across facts. This design
enhances the model's ability to edit multiple facts concurrently under practical
metrics, while also improving robustness against prefix perturbations and
simultaneous edits of same-subject samples. We elaborate on the details below.

\textbf{Key Embedding Preparation (\cref{method_overview} (a)).} Before
optimization, we extract the key embeddings corresponding to each knowledge item
that is scheduled for editing. For a given knowledge item $i$, we calculate the
cosine similarity between its key embedding $k_i$ and the key embeddings of all
other items. We then collect these similarity values into the set $P_k^{(i)} =
\{ P_k^{(i,j)} = \cos(k_i, k_j) \mid j \ne i \}$.

\textbf{Aligning Memory Embeddings with Key Embeddings (\cref{method_overview}
(b)).} For $N$ knowledge items, we separately optimize the target memory
embeddings to update factual associations. During the iterative optimization
process, we save every optimized residual embedding. When optimizing the target
memory for the $i$-th knowledge item, we compute the cosine similarity between
$r_i$ and all residual embeddings saved so far, and collect them as $P_r^{(i)} =
\{ P_r^{(i,j)} \mid j < i \}$. To promote alignment between key and residual
embeddings, we compute the KL divergence \cite{chen2020simple,
he2020momentum, sun2016deep} between $P_r^{(i)}$ and $\bar{P}_k^{(i)}$, where
$\bar{P}_k^{(i)} = \{ P_k^{(i,j)} \mid j < i \}$ denotes the subset of
$P_k^{(i)}$ corresponding to earlier items:  
\begin{equation}
    \label{kl_divergence}
    L_{\text{KL}}(i) = \text{KL}\!\left(P_r^{(i)} \,\|\, \bar{P}_k^{(i)}\right).
\end{equation}
Since KL divergence emphasizes distributional differences, we further strengthen
the alignment by selecting the top $M$ cosine similarities $\{P_k^{(i,j)}\}$
from $P_k^{(i)}$, and computing the mean squared error (MSE) loss between the
corresponding residual similarities $\{P_r^{(i,j)}\}$:  
\begin{equation}
    \label{mse_loss}
    L_{\text{MSE}}(i) = \frac{1}{M} \sum_{j=1}^M \big\| P_r^{(i,j)} - P_k^{(i,j)} \big\|^2.
\end{equation}


\textbf{Optimizing the Target Memory (\cref{method_overview} (c)).}
Our goal in this step is to compute the residual update vector $r_i$ for each
factual association $(s_i, rel_i, o_i)$ such that the model reliably predicts
the target object $o_i$ while preserving the alignment between the memory
embeddings and the key embeddings. To make the optimization procedure explicit,
we describe each component of the objective in \cref{optimize_m}. For each fact
$i$, let $h_i^L$ denote the hidden state at layer $L$ produced by the templated
prompt $tp(s_i, rel_i)$. Following prior
work~\citep{meng2022locating,meng2023memit}, we augment this prompt with a set
of $N_{\mathrm{FP}}$ randomly sampled prefixes
$\{f_j\}_{j=1}^{N_{\mathrm{FP}}}$, forming inputs $f_j \oplus tp(s_i, rel_i)$.
These prefixes encourage the model to learn more generalizable memory
representations. We write the forward pass of the model with the edited hidden
state as $G_{(h_i^L += r_i)}$, indicating that the hidden representation at
layer $L$ is perturbed by the update vector~$r_i$.

Given these definitions, we optimize $r_i$ by minimizing the following loss:
\begin{equation}
    \label{optimize_m}
    r_i = \underset{r_i}{\arg\min}\Bigg(
    \frac{1}{N_{\mathrm{FP}}} \sum_{j=1}^{N_{\mathrm{FP}}}
    -\log 
    \mathds{P}_{G(h_i^L += r_i)}\!\left[o_i \,\middle|\, f_j \oplus tp(s_i, rel_i)\right]
    + \lambda_{\mathrm{KL}} L_{\mathrm{KL}}(i)
    + \lambda_{\mathrm{MSE}} L_{\mathrm{MSE}}(i)
    \Bigg).
\end{equation}

Here, the first term encourages the model to predict the correct target object
$o_i$ under all sampled prefixes. The losses $L_{\mathrm{KL}}(i)$ and
$L_{\mathrm{MSE}}(i)$ ensure that the alignment between the memory embeddings
and the key embeddings is preserved, with $\lambda_{\mathrm{KL}}$ and
$\lambda_{\mathrm{MSE}}$ controlling their relative importance.

The full optimization procedure is detailed in \cref{appendix:full_algorithm}.
We justify our design of combining KL loss and MSE loss in
\cref{appendix:ablation}. As the optimization process is iterative, the editing
order of knowledge items may influence the performance of EAMET. We further
investigate the robustness of EAMET against different editing orders in
\cref{tab:editing_sequence}.

\section{Experiments}
\label{exps}
In this section, we empirically focus on evaluating the following research
questions (RQs). We first demonstrate the \emph{effectiveness} of EAMET in massive by
considering:
\begin{itemize}[leftmargin=2em, noitemsep, topsep=0pt]
    \item \textbf{RQ1.} Can EAMET generate more aligned embeddings for
    different knowledge items?
    \item \textbf{RQ2.} How does EAMET perform on massive editing tasks compared
    with baselines for various LLMs? Can it excel under the practical metric?
\end{itemize}
We then examine the \emph{robustness} of EAMET in two representative scenarios:
\begin{itemize}[leftmargin=2em, noitemsep, topsep=0pt]
    \item \textbf{RQ3.} How does EAMET perform when evaluating edited facts with
    prefixes?
    \item \textbf{RQ4.} How does EAMET perform when editing multiple facts of
    the same subject?
\end{itemize}

\subsection{Experiments Setup}
\label{main_text_exps_setup}

\textbf{Models, Datasets, and Baselines.}
We conduct extensive experiments on various LLMs, including
LLaMA2-7B~\citep{touvron2023llama2openfoundation},
LLaMA2-13B~\citep{touvron2023llama2openfoundation},
Falcon-7B~\citep{almazrouei2023falcon}, Qwen-2.5-7B~\citep{yang2024qwen2},
Deepseek-base-7B~\citep{bi2024deepseek}, and LLaMA3-8B~\citep{touvron2023llama}.
We provide additional evaluations on more LLMs in \cref{appendix:more_models}.
We consider a range of ME techniques as baselines: FT~\citep{zhu2020modifying},
MEND~\citep{mitchell2022fast}, ROME~\citep{meng2022locating},
MEMIT~\citep{meng2023memit}, PMET~\citep{10.1609/aaai.v38i17.29818}, and
ALPHAEDIT~\citep{fang2024alphaedit}. We demonstrate their performance on
CounterFact~\citep{meng2022locating}, ZsRE~\citep{levy2017zero}, and
Wiki-recent~\citep{zhang2024comprehensive}. We provide a full description in
\cref{appendix:datasets_baselines}.

\textbf{Evaluation Metrics.} Following previous work, we evaluate the
performance of ME techniques in terms of efficacy (Eff.), generalization (Gen.),
specificity (Spe.), and fluency (Flu.) for CounterFact and ZsRE datasets. For
Wiki-recent, we additionally evaluate the
portability~\citep{zhang2024comprehensive} (Por.) of edited models, which represents
the ability to address downstream tasks with edited knowledge. We propose to
evaluate the editing performance of ME techniques by requiring the edited models
to strictly examine whether explicit target objects are retrieved, as
demonstrated in \cref{intro_fig}. The editing efficacy is then defined as:
\begin{equation}
    \label{practical_metric}
    \text{Eff.} = \mathds{E}_i[o_i=\arg\mathop{\max}_{o} \mathds{P}_{f_\theta}(o\ |\ (s_i,rel_i))].
\end{equation}
When evaluating efficacy on the CounterFact and Wiki-recent datasets, and
generalization on CounterFact, we prepend each prompt with 10 distinct 5-token
prefixes. Full details of metrics are provided in \cref{appendix:metrics}. We
also provide the implementation details of EAMET in
\cref{implementation_details}.

\begin{wraptable}{!h}{0.55\textwidth}
    \vspace{-0.1in}
    \caption{Misalignment score comparison between different methods. Here,
    ``CF'' and ``ZS'' denote the CounterFact and ZsRE datasets, respectively.}
    \scalebox{0.8}{
      \begin{tabular}{c|cc|cc|cc}
      \toprule
      \multicolumn{1}{c}{\multirow{2}[2]{*}{\textbf{Model}}} & \multicolumn{2}{c}{\textbf{EAMET}}
      & \multicolumn{2}{c}{\textbf{MEMIT}} & \multicolumn{2}{c}{\textbf{PMET}} \\
      \cmidrule(lr){2-3}\cmidrule(lr){4-5}\cmidrule(lr){6-7}
      \multicolumn{1}{c}{} & \textbf{CF} & \multicolumn{1}{c}{\textbf{ZS}}
      & \textbf{CF} & \multicolumn{1}{c}{\textbf{ZS}} & \textbf{CF} & \multicolumn{1}{c}{\textbf{ZS}} \\
      \midrule
      LLaMA2-7B & \textbf{377}   & \textbf{165}   & 11506 & 22245 & 11475 & 11477 \\
      Qwen-7B & \textbf{374}   & \textbf{180}   & 18498 & 23699 & 18471 & 18463 \\
      Deepseek-7B & \textbf{520}   & \textbf{161}   & 12135 & 23241 & 12155 & 12046 \\
      Falcon-7B & \textbf{385}   & \textbf{181}   & 8564  & 17589 & 8602  & 8590 \\
      \bottomrule
      \end{tabular}
    }
    \label{tab:rq1}%
    \vspace{-0.1in}
\end{wraptable}

\subsection{Alignment of Retrieved Embeddings (RQ1)}
\textbf{\textit{Finding 1.} EAMET Promotes More Aligned Embeddings.} We compute
the summation of the misalignment score between the residual and key embeddings
for 10,000 facts edited by MEMIT, PMET, and EAMET under various LLMs. As shown
in \cref{tab:rq1}, the residual embeddings generated by EAMET are more aligned
with the key embeddings, while those produced by MEMIT and PMET are more likely
to cause inconsistency in the key and residual embeddings space. This
observation supports our hypothesis that EAMET encourages more aligned target
memory embeddings.
\subsection{Performance of Massive Editing (RQ2)}
\begin{table}[h!]
    \centering
    \vspace{-0.2in}
    \caption{Performance comparison of different editing methods on six LLMs over
    the Counterfact, Wiki-recent, and ZsRE benchmarks. We report the average
    value calculated over five evaluations.} 
    \scalebox{0.7}{
    \begin{tabular}{cc|cccc|cccc|ccc}
    \toprule
    \multicolumn{1}{c}{\multirow{2}{*}{\textbf{Model}}} &
    \multicolumn{1}{c}{\multirow{2}{*}{\textbf{Method}}} &
    \multicolumn{4}{c}{\textbf{Counterfact}} &
    \multicolumn{4}{c}{\textbf{Wiki-recent}} &
    \multicolumn{3}{c}{\textbf{ZsRE}} \\
    \cmidrule(lr){3-6}\cmidrule(lr){7-10}\cmidrule(lr){11-13}
    \multicolumn{1}{c}{} & \multicolumn{1}{c}{} & \textbf{Eff.}$\uparrow$ & \textbf{Gen.}$\uparrow$ & \textbf{Spe.}$\uparrow$ & \textbf{Flu.}$\uparrow$ & \textbf{Eff.}$\uparrow$ & \textbf{Por.}$\uparrow$ & \textbf{Loc.}$\uparrow$ & \textbf{Flu.}$\uparrow$ & \textbf{Eff.}$\uparrow$ & \textbf{Gen.}$\uparrow$ & \textbf{Spe.}$\uparrow$ \\
    \midrule
    \multirow{7}{*}{\rotatebox{90}{LLaMA2-7B}} 
     & FT & 0.29 & 0.23 & 77.43 & 490.34 & 7.23 & 41.61 & 36.52 & 491.83 & 5.30 & 4.31 & 14.69 \\
     & MEND & 0.23 & 0.31 & \textbf{78.55} & 307.26 & 0.00 & 34.67 & 37.46 & 269.52 & 0.00 & 0.00 & 0.50 \\
     & ROME & 0.00 & 0.00 & 50.73 & 467.76 & 76.73 & 49.31 & 51.51 & 497.53 & 37.29 & 6.86 & 10.27 \\
     & MEMIT & 24.95 & 22.68 & 63.84 & 506.69 & 34.75 & 44.93 & 46.72 & 504.18 & 76.63 & 64.06 & 15.57 \\
     & PMET & 74.22 & 46.45 & 72.47 & 507.10 & 81.84 & 51.11 & 53.16 & 497.49 & 77.29 & 71.40 & 16.54 \\
     & ALPHAEDIT & 0.51 & 0.53 & 51.14 & 501.63 & 0.07 & 35.34 & 37.48 & \textbf{527.83} & 44.26 & 35.83 & 12.65 \\
     & \textbf{EAMET} & \textbf{89.09} & \textbf{61.21} & 72.19 & \textbf{519.89} & \textbf{93.23} & \textbf{53.13} & \textbf{54.61} & 503.52 & \textbf{89.47} & \textbf{81.34} & \textbf{15.70} \\
    \midrule
    \multirow{7}{*}{\rotatebox{90}{Qwen2.5-7B}}
     & FT & 16.18 & 14.15 & 56.07 & 527.56 & 21.17 & 51.40 & 51.50 & \textbf{515.87} & 14.30 & 13.00 & 39.28 \\
     & MEND & 0.01 & 0.06 & 70.73 & 282.92 & 0.00 & 42.55 & 44.37 & 272.90 & 0.00 & 0.00 & 0.09 \\
     & ROME & 0.00 & 0.00 & 49.83 & 523.45 & 16.28 & 46.52 & 46.61 & 502.37 & 4.10 & 3.43 & 1.30 \\
     & MEMIT & 90.06 & 63.86 & 70.53 & 529.27 & 94.88 & 56.97 & \textbf{61.23} & 510.43 & 54.12 & 42.96 & 31.57 \\
     & PMET & 65.71 & 52.84 & 63.14 & 518.92 & 82.39 & \textbf{58.38} & 57.59 & 511.62 & 53.58 & 46.59 & 36.50 \\
     & ALPHAEDIT & 83.15 & 55.70 & 67.16 & 514.07 & 94.16 & 57.17 & 59.45 & 510.32 & 44.52 & 34.98 & 25.52 \\
     & \textbf{EAMET} & \textbf{90.49} & \textbf{64.37} & \textbf{72.18} & \textbf{536.67} & \textbf{95.61} & 57.46 & 60.28 & 509.06 & \textbf{91.03} & \textbf{84.80} & \textbf{41.20} \\
    \midrule
    \multirow{6}{*}{\rotatebox{90}{LLaMA2-13B}}
     & FT & 1.23 & 0.07 & 68.57 & 484.56 & 13.90 & 36.89 & 40.09 & 497.21 & 5.95 & 5.10 & 15.16 \\
     & ROME & 4.05 & 1.52 & 50.44 & 525.12 & 11.06 & 38.14 & 39.09 & 447.42 & 5.52 & 5.06 & 2.25 \\
     & MEMIT & 47.98 & 34.75 & 71.61 & 517.63 & 94.76 & 51.38 & 50.40 & 507.84 & 69.15 & 51.58 & 15.53 \\
     & PMET & 78.60 & 38.76 & \textbf{81.15} & 526.82 & 88.66 & 49.69 & 47.58 & 501.61 & 53.27 & 35.73 & 15.76 \\
     & ALPHAEDIT & 3.03 & 1.9 & 54.97 & 421.97 & 93.68 & 51.65 & 52.33 & \textbf{508.82} & 80.27 & 63.66 & 15.32 \\
     & \textbf{EAMET} & \textbf{92.85} & \textbf{60.08} & 77.51 & \textbf{530.78} & \textbf{95.88} & \textbf{52.08} & \textbf{53.43} & 504.06 & \textbf{87.09} & \textbf{74.58} & \textbf{15.90} \\
    \midrule
    \multirow{6}{*}{\rotatebox{90}{Falcon-7B}}
     & FT & 14.70 & 13.54 & 56.34 & 167.18 & 23.94 & 50.46 & 49.69 & 351.18 & 13.64 & 12.68 & 32.28 \\
     & ROME & 12.85 & 12.56 & 51.48 & 353.38 & 74.57 & 52.10 & 53.64 & \textbf{510.92} & 8.39 & 7.3 & 10.29 \\
     & MEMIT & 89.21 & 60.85 & 77.56 & 519.92 & 96.04 & 55.23 & 56.91 & 497.35 & 82.93 & 68.93 & 33.64 \\
     & PMET & 77.61 & 57.03 & 70.48 & 517.09 & 58.03 & 54.40 & 54.49 & 500.87 & 69.73 & 60.69 & 35.34 \\
     & ALPHAEDIT & 87.62 & 58.32 & 72.43 & 500.35 & 96.22 & 55.47 & 58.02 & 493.56 & 53.78 & 40.83 & 22.60 \\
     & \textbf{EAMET} & \textbf{92.37} & \textbf{63.91} & \textbf{78.94} & \textbf{528.98} & \textbf{96.94} & \textbf{57.08} & \textbf{58.58} & 507.56 & \textbf{92.38} & \textbf{81.15} & \textbf{36.71} \\
    \midrule 
    \multirow{6}{*}{\rotatebox{90}{Deepseek-7B}}
     & FT & 2.61 & 2.49 & \textbf{81.43} & \textbf{519.35} & 18.85 & 48.90 & 52.78 & 500.86 & 15.00 & 12.28 & 39.14 \\
     & ROME & 0.26 & 0.30 & 49.82 & 514.72 & 0.55 & 43.17 & 46.02 & 406.64 & 0.81 & 0.78 & 0.75 \\
     & MEMIT & 62.11 & 42.01 & 78.04 & 512.16 & 33.65 & 52.28 & 49.05 & 499.49 & 57.10 & 42.58 & 39.12 \\
     & PMET & 74.52 & 43.49 & 79.01 & 514.58 & 86.75 & \textbf{57.85} & 59.93 & 500.50 & 76.97 & 69.22 & 38.47 \\
     & ALPHAEDIT & 22.51 & 14.00 & 59.92 & 479.52 & 18.53 & 48.33 & 48.74 & 483.38 & 73.41 & 57.09 & 34.87 \\
     & \textbf{EAMET} & \textbf{89.74} & \textbf{59.98} & 77.73 & 513.93 & \textbf{97.15} & 56.43 & \textbf{60.45} & \textbf{501.09} & \textbf{87.27} & \textbf{70.02} & \textbf{39.87} \\
    \midrule
    \multirow{6}{*}{\rotatebox{90}{LLaMA3-8B}}
     & FT & 2.68 & 1.30 & 58.16 & 434.67 & 16.05 & 47.51 & 48.84 & 490.71 & 11.75 & 10.48 & 40.53 \\
     & ROME & 51.01 & 33.32 & 64.37 & 491.98 & 82.19 & 54.94 & 57.74 & 518.89 & 7.40 & 6.82 & 27.79 \\
     & MEMIT & 93.76 & 61.98 & 77.69 & 526.47 & 92.63 & 55.60 & 58.75 & 527.29 & 78.40 & 71.76 & 39.21 \\
     & PMET & 77.71 & 49.41 & 71.43 & 510.82 & 75.81 & 56.89 & 58.26 & 513.84 & 68.52 & 62.72 & 39.35 \\
     & ALPHAEDIT & 58.97 & 33.02 & \textbf{85.16} & \textbf{537.91} & 65.36 & 51.44 & 53.55 & 516.53 & 64.01 & 57.01 & 40.82 \\
     & \textbf{EAMET} & \textbf{93.87} & \textbf{63.74} & 79.07 & 533.30 & \textbf{94.36} & \textbf{57.88} & \textbf{59.48} & \textbf{528.23} & \textbf{85.68} & \textbf{81.34} & \textbf{42.39} \\
    \bottomrule
    \end{tabular}
    }
    \label{exp_massive_editing}
    \vspace{-0.1in}
\end{table}

We demonstrate the effectiveness of EAMET in massive editing tasks by comparing
it with baseline methods across six popular LLMs. Specifically, we
simultaneously edit 10{,}000 factual associations sampled from the CounterFact
and ZsRE datasets. For the Wiki-recent dataset, we modify all 1{,}266 knowledge
items. As shown in \cref{exp_massive_editing}, our key findings are as follows:

\begin{wraptable}{!h}{0.42\textwidth}
    \vspace{-0.2in}
    \caption{Performance comparison of different editing methods on Qwen2.5-7B,
    Falcon-7B, and LLaMA3-8B with 15{,}000 edits from the CounterFact benchmark.
    }
    \setlength{\tabcolsep}{4pt}
    \scalebox{0.7}{
    \begin{tabular}{cc|cccc}
    \toprule
    \multicolumn{1}{c}{\multirow{2}{*}{\textbf{Model}}} &
    \multicolumn{1}{c}{\multirow{2}{*}{\textbf{Method}}} &
    \multicolumn{4}{c}{\textbf{Counterfact (15000)}} \\
    \cmidrule(lr){3-6}
    \multicolumn{1}{c}{} & \multicolumn{1}{c}{} & \textbf{Eff.}$\uparrow$ & \textbf{Gen.}$\uparrow$ & \textbf{Spe.}$\uparrow$ & \textbf{Flu.}$\uparrow$ \\
    \midrule
    \multirow{2}{*}{Qwen2.5-7B}
        & MEMIT & 77.46 & 54.34 & 66.23 & 514.81 \\
        & \textbf{EAMET} & \textbf{83.66} & \textbf{55.31} & \textbf{69.49} & \textbf{528.28} \\
    \multirow{2}{*}{Falcon-7B}
        & MEMIT & 84.60 & 56.13 & \textbf{75.51} & 513.82 \\
        & \textbf{EAMET} & \textbf{89.55} & \textbf{61.00} & 68.44 & \textbf{516.91} \\
    \multirow{2}{*}{LLaMA3-8B}
        & MEMIT & 87.58 & 54.76 & 72.65 & 514.07 \\
        & \textbf{EAMET} & \textbf{91.22} & \textbf{62.24} & \textbf{73.43} & \textbf{531.76} \\
    \bottomrule
    \end{tabular}
    }
    \label{exp_massive_editing_more}
    \vspace{-0.1in}
\end{wraptable}

\textbf{\textit{Finding 2.} EAMET Consistently Achieves Superior Editing
Performance Across All Datasets and Model Architectures.} Across all evaluated
datasets, EAMET demonstrates the highest levels of editing efficacy and
generalization. On the CounterFact dataset, it consistently outperforms other
methods, particularly on base models such as LLaMA2-7B, LLaMA2-13B, and
Deepseek-7B. For example, EAMET achieves 89.09\% efficacy and 61.21\%
generalization on LLaMA2-7B, outperforming the second-best method (PMET) by 15\%
on both metrics. The gap widens further compared to MEMIT, with improvements of
65\% in efficacy and 39\% in generalization. Even on more advanced models such
as Qwen2.5-7B, Falcon-7B, and LLaMA3-8B, EAMET consistently surpasses all
baselines. Furthermore, its advantage becomes more pronounced at larger editing
scales. As shown in \cref{exp_massive_editing_more}, when editing 15{,}000
knowledge items on Qwen2.5-7B, EAMET achieves 83.66\% efficacy, demonstrating a
10\% improvement over MEMIT. We additionally report the superior performance of
EAMET across diverse semantic scenarios in \cref{appendix:semantics}.

\textbf{\textit{Finding 3.} EAMET Preserves the General Abilities of the Edited
models.} In addition to achieving state-of-the-art editing performance, EAMET
does not impair the base model's fluency or reasoning abilities. Across all
datasets, EAMET consistently attains among the highest specificity and fluency
scores. Notably, on the Wiki-recent dataset, EAMET achieves the best portability
performance on most base models, indicating that the edited models retain their
ability to reason about downstream knowledge related to the edited facts. We
also evaluate the general abilities of edited models on GLUE
\citep{wang-etal-2018-glue} and find that EAMET yields minimal deviation from
pre-edit performance (\cref{appendix:glue_benchmarks}).
\begin{table}[h!]
    \centering
    \caption{Impact of editing sequence on EAMET's performance on Counterfact and ZsRE datasets.}
    \small
    \scalebox{1}{
    \begin{tabular}{l|cccc|ccc}
    \toprule
    \multicolumn{1}{c}{\multirow{2}{*}{\textbf{Method}}} &
    \multicolumn{4}{c}{\textbf{Counterfact}} &
    \multicolumn{3}{c}{\textbf{ZsRE}} \\
    \cmidrule(lr){2-5}\cmidrule(lr){6-8}
    \multicolumn{1}{c}{} & \textbf{Eff.}$\uparrow$ & \textbf{Gen.}$\uparrow$ & \textbf{Spe.}$\uparrow$ & \textbf{Flu.}$\uparrow$ & \textbf{Eff.}$\uparrow$ & \textbf{Gen.}$\uparrow$ & \textbf{Spe.}$\uparrow$ \\
    \midrule
    EAMET (original sequence)& 89.09 & 61.21 & 72.19 & 519.06 & 89.47 & 81.34 & 15.70 \\
    -- random shuffle (seed=0) & 88.21 & 60.79 & 71.84 & 519.21 & 87.63 & 77.42 & 15.56 \\
    -- random shuffle (seed=1) & 89.11 & 60.78 & 72.03 & 518.84 & 86.99 & 76.08 & 15.58 \\
    -- random shuffle (seed=2) & 88.91 & 59.38 & 72.34 & 518.23 & 87.56 & 77.47 & 15.59 \\
    \bottomrule
    \end{tabular}
    }
    \label{tab:editing_sequence}
\end{table}

As EAMET preserves previously optimized residual embeddings when updating new
knowledge items, the editing sequence could potentially affect its performance.
To assess this, we examine EAMET's robustness under different editing orders on
the Counterfact and ZsRE datasets. In Counterfact, all knowledge items have
distinct subjects, whereas in ZsRE some items share the same subject and are
adjacent in the original order. We therefore randomly shuffle the order of
10,000 items three times and report the average performance, alongside the
original sequence as a reference.

\noindent \textbf{\textit{Finding 4.} EAMET is Robust to Editing
Sequence.} As shown in \cref{tab:editing_sequence}, EAMET's performance remains
stable across editing orders. On Counterfact, random shuffles produce only
negligible variations in efficacy, generalization, and specificity. On ZsRE,
editing efficacy shows a slight decline of about 2\%, likely due to the
neighborhood structure of items sharing the same subject in the original
sequence. Overall, these results suggest that EAMET is largely insensitive to
editing order, demonstrating strong robustness to sequence variations.

\subsection{Robustness against long prefixes (RQ3)}
\begin{figure}[!h]
    \vspace{-0.1in}
    \begin{center}
    \scalebox{1}{
    \centerline{\includegraphics[width=\columnwidth]{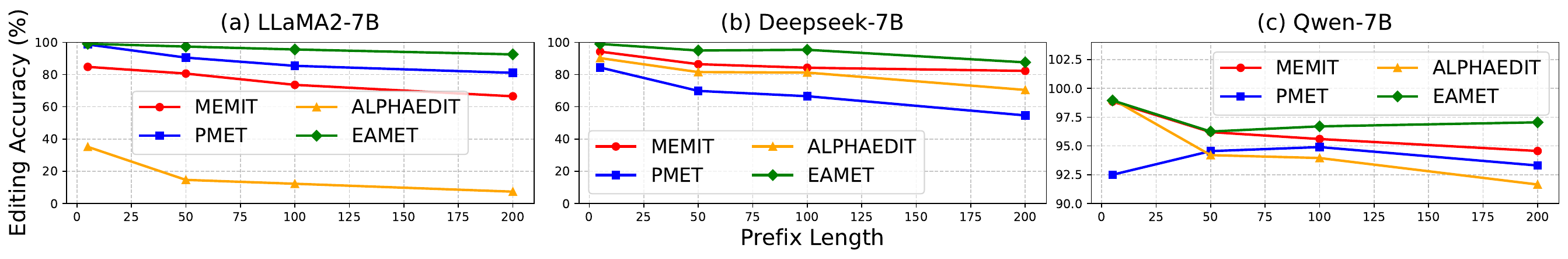}}
    }
    \caption{Editing performance of different methods across varying prefix lengths.}
    \label{prefix_length_counterfact}
    \end{center}
    \vspace{-0.1in}
\end{figure}

We evaluate the robustness of editing methods when edited facts are
preceded by varying numbers of tokens. Specifically, we modify 200 facts from
the CounterFact dataset in LLaMA2-7B, Deepseek-7B, and Qwen2.5-7B. During
evaluation, we prepend prefixes of 5, 50, 100, and 200 tokens to them.

\textbf{\textit{Finding 5.} EAMET Remains Effective When Edits Are Preceded by
Long Prefixes.} In \cref{prefix_length_counterfact}, EAMET achieves the highest
editing efficacy across all models, with at most a 7\% drop at 200-token
prefixes. In contrast, MEMIT suffers a much larger decline, from 84.75\% to
66.50\% on LLaMA2-7B and from 94.2\% to 82.25\% on DeepSeek-7B. Notably, all
methods demonstrate strong robustness on Qwen2.5-7B, consistent with our earlier
observation that Qwen2.5-7B is more suitable for robust batch editing.
Nevertheless, EAMET exhibits the smallest efficacy drop (only 1.9\%) when the
prefix increases to 200 tokens, which is half that of the second-best method
(MEMIT).

\subsection{Robustness under Multiple Edits of the Same Subject (RQ4)}
\begin{figure}[!h]
    \vspace{-0.1in}
    \begin{center}
    \scalebox{1}{
    \centerline{\includegraphics[width=\columnwidth]{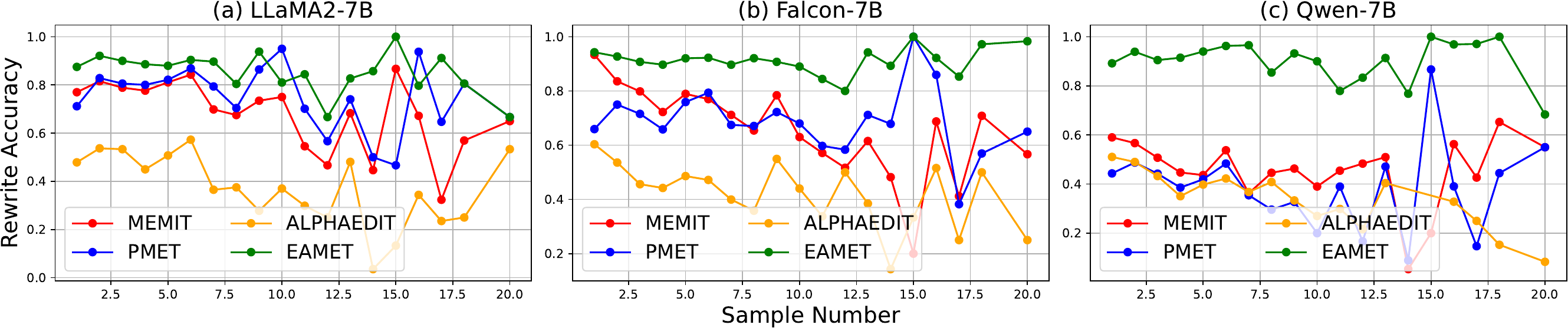}}
    }
    \caption{Editing performance of different methods across varying numbers of facts per subject.}
    \label{editing_for_same_subject}
    \end{center}
    \vspace{-0.1in}
\end{figure}

We evaluate the robustness of editing methods when multiple facts concerning the
same subject are edited simultaneously. Specifically, we simultaneously edit
10,000 facts from ZsRE dataset, and only evaluate samples whose subject is
associated with multiple facts. We group subjects according to the number of
associated samples and examine how rewrite accuracy varies with this number.
Experiments are conducted on LLaMA2-7B, DeepSeek-7B, and Qwen2.5-7B.

\textbf{\textit{Finding 6.} EAMET Remains Effective When Multiple Facts of the
Same Subject Are Edited Simultaneously.} \cref{editing_for_same_subject} shows
that EAMET consistently achieves the highest editing efficacy across nearly all
settings. Its performance remains stable when editing multiple samples
associated with the same subject. In contrast, other methods exhibit a clear
decline in efficacy as the number of facts per subject increases, which
ultimately results in degraded performance on the overall massive editing task.

\section{Conclusion}
\label{conclusion}
In this paper, we propose EAMET, a novel model editing method that enables
stronger and more robust massive editing across various models and datasets. We
first identify that the failures of existing methods in both effectiveness and
robustness of massive editing stem from misalignment between the space of key
and residual embeddings. EAMET addresses this issue by progressively aligning
the key and residual embedding space when optimizing target memory for each
fact. The aligned embeddings increase both the capacity and robustness of
massive editing. Extensive experiments on multiple base LLMs, including LLaMA2,
LLaMA3, Deepseek, and Qwen, demonstrate that EAMET significantly outperforms
existing methods in editing performance and robustness.

\section*{Acknowledgments}
The HKUST authors were supported in part by a RGC GRF grant under the contract
16214723, research fund provided by CMHK and ZTE, and a HKUST Bridge The Gap
fund BGF.001.2025. We are grateful to the anonymous reviewers for their valuable
comments. We thank HKUST Fok Ying Tung Research Institute and National
Supercomputing Center in Guangzhou Nansha Sub-center for computational
resources.

\section*{Ethics Statement}
This work does not raise any specific ethical concerns. Its primary goal is to
advance research on model editing and to offer a new perspective on this
problem.

\section*{Reproducibility Statement}
All experiments in this paper are reproducible. The code and datasets are
publicly available at the GitHub repository:
\url{https://github.com/ybdai7/EAMET-massive-editing}.

\bibliography{iclr2026_conference}

@inproceedings{mitchell2022memory,
  title={Memory-based model editing at scale},
  author={Mitchell, Eric and Lin, Charles and Bosselut, Antoine and Manning, Christopher D and Finn, Chelsea},
  booktitle={International Conference on Machine Learning},
  pages={15817--15831},
  year={2022},
  organization={PMLR}
}

@inproceedings{zheng-etal-2023-edit,
    title = "Can We Edit Factual Knowledge by In-Context Learning?",
    author = "Zheng, Ce  and
      Li, Lei  and
      Dong, Qingxiu  and
      Fan, Yuxuan  and
      Wu, Zhiyong  and
      Xu, Jingjing  and
      Chang, Baobao",
    editor = "Bouamor, Houda  and
      Pino, Juan  and
      Bali, Kalika",
    booktitle = "Proceedings of the 2023 Conference on Empirical Methods in Natural Language Processing",
    month = dec,
    year = "2023",
    address = "Singapore",
    publisher = "Association for Computational Linguistics",
    url = "https://aclanthology.org/2023.emnlp-main.296/",
    doi = "10.18653/v1/2023.emnlp-main.296",
    pages = "4862--4876"
}

@inproceedings{NEURIPS2023_95b6e2ff,
	author = {Hartvigsen, Tom and Sankaranarayanan, Swami and Palangi, Hamid and Kim, Yoon and Ghassemi, Marzyeh},
	booktitle = {Advances in Neural Information Processing Systems},
	editor = {A. Oh and T. Naumann and A. Globerson and K. Saenko and M. Hardt and S. Levine},
	pages = {47934--47959},
	publisher = {Curran Associates, Inc.},
	title = {Aging with GRACE: Lifelong Model Editing with Discrete Key-Value Adaptors},
	volume = {36},
	year = {2023},
}

@inproceedings{yu2024melo,
  title={Melo: Enhancing model editing with neuron-indexed dynamic lora},
  author={Yu, Lang and Chen, Qin and Zhou, Jie and He, Liang},
  booktitle={Proceedings of the AAAI Conference on Artificial Intelligence},
  volume={38},
  number={17},
  pages={19449--19457},
  year={2024}
}

@inproceedings{
mitchell2022fast,
title={Fast Model Editing at Scale},
author={Eric Mitchell and Charles Lin and Antoine Bosselut and Chelsea Finn and Christopher D Manning},
booktitle={International Conference on Learning Representations},
year={2022},
url={https://openreview.net/forum?id=0DcZxeWfOPt}
}

@article{meng2022locating,
  title={Locating and Editing Factual Associations in {GPT}},
  author={Kevin Meng and David Bau and Alex Andonian and Yonatan Belinkov},
  journal={Advances in Neural Information Processing Systems},
  volume={36},
  year={2022},
  note={arXiv:2202.05262}
}

@article{meng2023memit,
  title={Mass Editing Memory in a Transformer},
  author={Kevin Meng and Sen Sharma, Arnab and Alex Andonian and Yonatan Belinkov and David Bau},
  journal={The Eleventh International Conference on Learning Representations (ICLR)},
  year={2023}
}

@inproceedings{10.1609/aaai.v38i17.29818,
author = {Li, Xiaopeng and Li, Shasha and Song, Shezheng and Yang, Jing and Ma, Jun and Yu, Jie},
title = {PMET: precise model editing in a transformer},
year = {2025},
isbn = {978-1-57735-887-9},
publisher = {AAAI Press},
url = {https://doi.org/10.1609/aaai.v38i17.29818},
doi = {10.1609/aaai.v38i17.29818},
booktitle = {Proceedings of the Thirty-Eighth AAAI Conference on Artificial Intelligence and Thirty-Sixth Conference on Innovative Applications of Artificial Intelligence and Fourteenth Symposium on Educational Advances in Artificial Intelligence},
articleno = {2070},
numpages = {9},
series = {AAAI'24/IAAI'24/EAAI'24}
}

@inproceedings{geva-etal-2021-transformer,
    title = "Transformer Feed-Forward Layers Are Key-Value Memories",
    author = "Geva, Mor  and
      Schuster, Roei  and
      Berant, Jonathan  and
      Levy, Omer",
    editor = "Moens, Marie-Francine  and
      Huang, Xuanjing  and
      Specia, Lucia  and
      Yih, Scott Wen-tau",
    booktitle = "Proceedings of the 2021 Conference on Empirical Methods in Natural Language Processing",
    month = nov,
    year = "2021",
    address = "Online and Punta Cana, Dominican Republic",
    publisher = "Association for Computational Linguistics",
    url = "https://aclanthology.org/2021.emnlp-main.446/",
    doi = "10.18653/v1/2021.emnlp-main.446",
    pages = "5484--5495"
}

@inproceedings{geva-etal-2022-transformer,
    title = "Transformer Feed-Forward Layers Build Predictions by Promoting Concepts in the Vocabulary Space",
    author = "Geva, Mor  and
      Caciularu, Avi  and
      Wang, Kevin  and
      Goldberg, Yoav",
    editor = "Goldberg, Yoav  and
      Kozareva, Zornitsa  and
      Zhang, Yue",
    booktitle = "Proceedings of the 2022 Conference on Empirical Methods in Natural Language Processing",
    month = dec,
    year = "2022",
    address = "Abu Dhabi, United Arab Emirates",
    publisher = "Association for Computational Linguistics",
    url = "https://aclanthology.org/2022.emnlp-main.3/",
    doi = "10.18653/v1/2022.emnlp-main.3",
    pages = "30--45"
}

@inproceedings{dai-etal-2022-knowledge,
    title = "Knowledge Neurons in Pretrained Transformers",
    author = "Dai, Damai  and
      Dong, Li  and
      Hao, Yaru  and
      Sui, Zhifang  and
      Chang, Baobao  and
      Wei, Furu",
    editor = "Muresan, Smaranda  and
      Nakov, Preslav  and
      Villavicencio, Aline",
    booktitle = "Proceedings of the 60th Annual Meeting of the Association for Computational Linguistics (Volume 1: Long Papers)",
    month = may,
    year = "2022",
    address = "Dublin, Ireland",
    publisher = "Association for Computational Linguistics",
    url = "https://aclanthology.org/2022.acl-long.581/",
    doi = "10.18653/v1/2022.acl-long.581",
    pages = "8493--8502"
}

@inproceedings{yang-etal-2024-butterfly,
    title = "The Butterfly Effect of Model Editing: Few Edits Can Trigger Large Language Models Collapse",
    author = "Yang, Wanli  and
      Sun, Fei  and
      Ma, Xinyu  and
      Liu, Xun  and
      Yin, Dawei  and
      Cheng, Xueqi",
    editor = "Ku, Lun-Wei  and
      Martins, Andre  and
      Srikumar, Vivek",
    booktitle = "Findings of the Association for Computational Linguistics: ACL 2024",
    month = aug,
    year = "2024",
    address = "Bangkok, Thailand",
    publisher = "Association for Computational Linguistics",
    url = "https://aclanthology.org/2024.findings-acl.322/",
    doi = "10.18653/v1/2024.findings-acl.322",
    pages = "5419--5437"
}

@misc{ma2024robustnesseditinglargelanguage,
      title={On the Robustness of Editing Large Language Models}, 
      author={Xinbei Ma and Tianjie Ju and Jiyang Qiu and Zhuosheng Zhang and Hai Zhao and Lifeng Liu and Yulong Wang},
      year={2024},
      eprint={2402.05827},
      archivePrefix={arXiv},
      primaryClass={cs.CL},
      url={https://arxiv.org/abs/2402.05827}, 
}

@inproceedings{
li2024unveiling,
title={Unveiling the Pitfalls of Knowledge Editing for Large Language Models},
author={Zhoubo Li and Ningyu Zhang and Yunzhi Yao and Mengru Wang and Xi Chen and Huajun Chen},
booktitle={The Twelfth International Conference on Learning Representations},
year={2024},
url={https://openreview.net/forum?id=fNktD3ib16}
}

@misc{touvron2023llama2openfoundation,
      title={Llama 2: Open Foundation and Fine-Tuned Chat Models}, 
      author={Hugo Touvron and Louis Martin and Kevin Stone and Peter Albert and Amjad Almahairi and Yasmine Babaei and Nikolay Bashlykov and Soumya Batra and Prajjwal Bhargava and Shruti Bhosale and Dan Bikel and Lukas Blecher and Cristian Canton Ferrer and Moya Chen and Guillem Cucurull and David Esiobu and Jude Fernandes and Jeremy Fu and Wenyin Fu and Brian Fuller and Cynthia Gao and Vedanuj Goswami and Naman Goyal and Anthony Hartshorn and Saghar Hosseini and Rui Hou and Hakan Inan and Marcin Kardas and Viktor Kerkez and Madian Khabsa and Isabel Kloumann and Artem Korenev and Punit Singh Koura and Marie-Anne Lachaux and Thibaut Lavril and Jenya Lee and Diana Liskovich and Yinghai Lu and Yuning Mao and Xavier Martinet and Todor Mihaylov and Pushkar Mishra and Igor Molybog and Yixin Nie and Andrew Poulton and Jeremy Reizenstein and Rashi Rungta and Kalyan Saladi and Alan Schelten and Ruan Silva and Eric Michael Smith and Ranjan Subramanian and Xiaoqing Ellen Tan and Binh Tang and Ross Taylor and Adina Williams and Jian Xiang Kuan and Puxin Xu and Zheng Yan and Iliyan Zarov and Yuchen Zhang and Angela Fan and Melanie Kambadur and Sharan Narang and Aurelien Rodriguez and Robert Stojnic and Sergey Edunov and Thomas Scialom},
      year={2023},
      eprint={2307.09288},
      archivePrefix={arXiv},
      primaryClass={cs.CL},
      url={https://arxiv.org/abs/2307.09288}, 
}

@article{brown2020language,
  title={Language models are few-shot learners},
  author={Brown, Tom and Mann, Benjamin and Ryder, Nick and Subbiah, Melanie and Kaplan, Jared D and Dhariwal, Prafulla and Neelakantan, Arvind and Shyam, Pranav and Sastry, Girish and Askell, Amanda and others},
  journal={Advances in neural information processing systems},
  volume={33},
  pages={1877--1901},
  year={2020}
}

@article{touvron2023llama,
  title={Llama 2: Open foundation and fine-tuned chat models},
  author={Touvron, Hugo and Martin, Louis and Stone, Kevin and Albert, Peter and Almahairi, Amjad and Babaei, Yasmine and Bashlykov, Nikolay and Batra, Soumya and Bhargava, Prajjwal and Bhosale, Shruti and others},
  journal={arXiv preprint arXiv:2307.09288},
  year={2023}
}

@article{yang2024qwen2,
  title={Qwen2. 5 technical report},
  author={Yang, An and Yang, Baosong and Zhang, Beichen and Hui, Binyuan and Zheng, Bo and Yu, Bowen and Li, Chengyuan and Liu, Dayiheng and Huang, Fei and Wei, Haoran and others},
  journal={arXiv preprint arXiv:2412.15115},
  year={2024}
}

@article{bi2024deepseek,
  title={Deepseek llm: Scaling open-source language models with longtermism},
  author={Bi, Xiao and Chen, Deli and Chen, Guanting and Chen, Shanhuang and Dai, Damai and Deng, Chengqi and Ding, Honghui and Dong, Kai and Du, Qiushi and Fu, Zhe and others},
  journal={arXiv preprint arXiv:2401.02954},
  year={2024}
}

@article{vykopal2023disinformation,
  title={Disinformation capabilities of large language models},
  author={Vykopal, Ivan and Pikuliak, Mat{\'u}{\v{s}} and Srba, Ivan and Moro, Robert and Macko, Dominik and Bielikova, Maria},
  journal={arXiv preprint arXiv:2311.08838},
  year={2023}
}

@article{huang2025survey,
  title={A survey on hallucination in large language models: Principles, taxonomy, challenges, and open questions},
  author={Huang, Lei and Yu, Weijiang and Ma, Weitao and Zhong, Weihong and Feng, Zhangyin and Wang, Haotian and Chen, Qianglong and Peng, Weihua and Feng, Xiaocheng and Qin, Bing and others},
  journal={ACM Transactions on Information Systems},
  volume={43},
  number={2},
  pages={1--55},
  year={2025},
  publisher={ACM New York, NY}
}

@article{xu2024hallucination,
  title={Hallucination is inevitable: An innate limitation of large language models},
  author={Xu, Ziwei and Jain, Sanjay and Kankanhalli, Mohan},
  journal={arXiv preprint arXiv:2401.11817},
  year={2024}
}

@article{achiam2023gpt,
  title={Gpt-4 technical report},
  author={Achiam, Josh and Adler, Steven and Agarwal, Sandhini and Ahmad, Lama and Akkaya, Ilge and Aleman, Florencia Leoni and Almeida, Diogo and Altenschmidt, Janko and Altman, Sam and Anadkat, Shyamal and others},
  journal={arXiv preprint arXiv:2303.08774},
  year={2023}
}

@article{team2025gemma,
  title={Gemma 3 technical report},
  author={Team, Gemma and Kamath, Aishwarya and Ferret, Johan and Pathak, Shreya and Vieillard, Nino and Merhej, Ramona and Perrin, Sarah and Matejovicova, Tatiana and Ram{\'e}, Alexandre and Rivi{\`e}re, Morgane and others},
  journal={arXiv preprint arXiv:2503.19786},
  year={2025}
}

@article{fang2024alphaedit,
  title={Alphaedit: Null-space constrained knowledge editing for language models},
  author={Fang, Junfeng and Jiang, Houcheng and Wang, Kun and Ma, Yunshan and Jie, Shi and Wang, Xiang and He, Xiangnan and Chua, Tat-Seng},
  journal={arXiv preprint arXiv:2410.02355},
  year={2024}
}

@article{petroni2019language,
  title={Language models as knowledge bases?},
  author={Petroni, Fabio and Rockt{\"a}schel, Tim and Lewis, Patrick and Bakhtin, Anton and Wu, Yuxiang and Miller, Alexander H and Riedel, Sebastian},
  journal={arXiv preprint arXiv:1909.01066},
  year={2019}
}

@article{roberts2020much,
  title={How much knowledge can you pack into the parameters of a language model?},
  author={Roberts, Adam and Raffel, Colin and Shazeer, Noam},
  journal={arXiv preprint arXiv:2002.08910},
  year={2020}
}

@article{jiang2020can,
  title={How can we know what language models know?},
  author={Jiang, Zhengbao and Xu, Frank F and Araki, Jun and Neubig, Graham},
  journal={Transactions of the Association for Computational Linguistics},
  volume={8},
  pages={423--438},
  year={2020},
  publisher={MIT Press One Rogers Street, Cambridge, MA 02142-1209, USA journals-info~…}
}

@article{shin2020autoprompt,
  title={Autoprompt: Eliciting knowledge from language models with automatically generated prompts},
  author={Shin, Taylor and Razeghi, Yasaman and Logan IV, Robert L and Wallace, Eric and Singh, Sameer},
  journal={arXiv preprint arXiv:2010.15980},
  year={2020}
}

@article{levy2017zero,
  title={Zero-shot relation extraction via reading comprehension},
  author={Levy, Omer and Seo, Minjoon and Choi, Eunsol and Zettlemoyer, Luke},
  journal={arXiv preprint arXiv:1706.04115},
  year={2017}
}

@article{zhang2024comprehensive,
  title={A comprehensive study of knowledge editing for large language models},
  author={Zhang, Ningyu and Yao, Yunzhi and Tian, Bozhong and Wang, Peng and Deng, Shumin and Wang, Mengru and Xi, Zekun and Mao, Shengyu and Zhang, Jintian and Ni, Yuansheng and others},
  journal={arXiv preprint arXiv:2401.01286},
  year={2024}
}

@article{almazrouei2023falcon,
  title={The falcon series of open language models},
  author={Almazrouei, Ebtesam and Alobeidli, Hamza and Alshamsi, Abdulaziz and Cappelli, Alessandro and Cojocaru, Ruxandra and Debbah, M{\'e}rouane and Goffinet, {\'E}tienne and Hesslow, Daniel and Launay, Julien and Malartic, Quentin and others},
  journal={arXiv preprint arXiv:2311.16867},
  year={2023}
}

@inproceedings{ju-etal-2024-investigating,
    title = "Investigating Multi-Hop Factual Shortcuts in Knowledge Editing of Large Language Models",
    author = "Ju, Tianjie  and
      Chen, Yijin  and
      Yuan, Xinwei  and
      Zhang, Zhuosheng  and
      Du, Wei  and
      Zheng, Yubin  and
      Liu, Gongshen",
    editor = "Ku, Lun-Wei  and
      Martins, Andre  and
      Srikumar, Vivek",
    booktitle = "Proceedings of the 62nd Annual Meeting of the Association for Computational Linguistics (Volume 1: Long Papers)",
    month = aug,
    year = "2024",
    address = "Bangkok, Thailand",
    publisher = "Association for Computational Linguistics",
    url = "https://aclanthology.org/2024.acl-long.486/",
    doi = "10.18653/v1/2024.acl-long.486",
    pages = "8987--9001"
}

@inproceedings{gu-etal-2024-pokemqa,
    title = "{P}oke{MQA}: Programmable knowledge editing for Multi-hop Question Answering",
    author = "Gu, Hengrui  and
      Zhou, Kaixiong  and
      Han, Xiaotian  and
      Liu, Ninghao  and
      Wang, Ruobing  and
      Wang, Xin",
    editor = "Ku, Lun-Wei  and
      Martins, Andre  and
      Srikumar, Vivek",
    booktitle = "Proceedings of the 62nd Annual Meeting of the Association for Computational Linguistics (Volume 1: Long Papers)",
    month = aug,
    year = "2024",
    address = "Bangkok, Thailand",
    publisher = "Association for Computational Linguistics",
    url = "https://aclanthology.org/2024.acl-long.438/",
    doi = "10.18653/v1/2024.acl-long.438",
    pages = "8069--8083"
}

@article{team2024gemma,
  title={Gemma: Open models based on gemini research and technology},
  author={Team, Gemma and Mesnard, Thomas and Hardin, Cassidy and Dadashi, Robert and Bhupatiraju, Surya and Pathak, Shreya and Sifre, Laurent and Rivi{\`e}re, Morgane and Kale, Mihir Sanjay and Love, Juliette and others},
  journal={arXiv preprint arXiv:2403.08295},
  year={2024}
}

@article{li2023textbooks,
  title={Textbooks are all you need ii: phi-1.5 technical report},
  author={Li, Yuanzhi and Bubeck, S{\'e}bastien and Eldan, Ronen and Del Giorno, Allie and Gunasekar, Suriya and Lee, Yin Tat},
  journal={arXiv preprint arXiv:2309.05463},
  year={2023}
}

@article{zhu2020modifying,
  title={Modifying memories in transformer models},
  author={Zhu, Chen and Rawat, Ankit Singh and Zaheer, Manzil and Bhojanapalli, Srinadh and Li, Daliang and Yu, Felix and Kumar, Sanjiv},
  journal={arXiv preprint arXiv:2012.00363},
  year={2020}
}

@inproceedings{NEURIPS2024_d74033a2,
	author = {Pramanick, Shraman and Chellappa, Rama and Venugopalan, Subhashini},
	booktitle = {Advances in Neural Information Processing Systems},
	editor = {A. Globerson and L. Mackey and D. Belgrave and A. Fan and U. Paquet and J. Tomczak and C. Zhang},
	pages = {118807--118833},
	publisher = {Curran Associates, Inc.},
	title = {SPIQA: A Dataset for Multimodal Question Answering on Scientific Papers},
	url = {https://proceedings.neurips.cc/paper_files/paper/2024/file/d74033a247989e8f6f3bf9e0c9629fb5-Paper-Datasets_and_Benchmarks_Track.pdf},
	volume = {37},
	year = {2024}}

@inproceedings{NEURIPS2024_1568882b,
	author = {Romero, David and Lyu, Chenyang and Wibowo, Haryo Akbarianto and Lynn, Teresa and Hamed, Injy and Kishore, Aditya Nanda and Mandal, Aishik and Dragonetti, Alina and Abzaliev, Artem and Tonja, Atnafu Lambebo and Balcha, Bontu Fufa and Whitehouse, Chenxi and Salamea, Christian and Velasco, Dan John and Adelani, David Ifeoluwa and Le Meur, David and Villa-Cueva, Emilio and Koto, Fajri and Farooqui, Fauzan and Belcavello, Frederico and Batnasan, Ganzorig and Vallejo, Gisela and Caulfield, Grainne and Ivetta, Guido and Song, Haiyue and Ademtew, Henok Biadglign and Maina, Hern\'{a}n and Lovenia, Holy and Azime, Israel Abebe and Cruz, Jan Christian Blaise and Gala, Jay and Geng, Jiahui and Ortiz-Barajas, Jesus-German and Baek, Jinheon and Dunstan, Jocelyn and Alemany, Laura Alonso and Nagasinghe, Kumaranage Ravindu Yasas and Benotti, Luciana and D\textquotesingle Haro, Luis Fernando and Viridiano, Marcelo and Estecha-Garitagoitia, Marcos and Cabrera, Maria Camila Buitrago and Rodr\'{\i}guez-Cantelar, Mario and Jouitteau, M\'{e}lanie and Mihaylov, Mihail and Etori, Naome and Imam, Mohamed Fazli Mohamed and Adilazuarda, Muhammad Farid and Gochoo, Munkhjargal and Otgonbold, Munkh-Erdene and Niyomugisha, Olivier and Silva, Paula M\'{o}nica and Chitale, Pranjal and Dabre, Raj and Chevi, Rendi and Zhang, Ruochen and Diandaru, Ryandito and Cahyawijaya, Samuel and G\'{o}ngora, Santiago and Jeong, Soyeong and Purkayastha, Sukannya and Kuribayashi, Tatsuki and Clifford, Teresa and Jayakumar, Thanmay and Torrent, Tiago Timponi and Ehsan, Toqeer and Araujo, Vladimir and Kementchedjhieva, Yova and Burzo, Zara and Lim, Zheng Wei and Yong, Zheng Xin and Ignat, Oana and Nwatu, Joan and Mihalcea, Rada and Solorio, Thamar and Aji, Alham Fikri},
	booktitle = {Advances in Neural Information Processing Systems},
	editor = {A. Globerson and L. Mackey and D. Belgrave and A. Fan and U. Paquet and J. Tomczak and C. Zhang},
	pages = {11479--11505},
	publisher = {Curran Associates, Inc.},
	title = {CVQA: Culturally-diverse Multilingual Visual Question Answering Benchmark},
	url = {https://proceedings.neurips.cc/paper_files/paper/2024/file/1568882ba1a50316e87852542523739c-Paper-Datasets_and_Benchmarks_Track.pdf},
	volume = {37},
	year = {2024}}

@book{horn1985matrix,
  title={Matrix Analysis},
  author={Horn, Roger A and Johnson, Charles R},
  year={1985},
  publisher={Cambridge University Press}
}

@inproceedings{wang-etal-2018-glue,
    title = "{GLUE}: A Multi-Task Benchmark and Analysis Platform for Natural Language Understanding",
    author = "Wang, Alex  and
      Singh, Amanpreet  and
      Michael, Julian  and
      Hill, Felix  and
      Levy, Omer  and
      Bowman, Samuel",
    editor = "Linzen, Tal  and
      Chrupa{\l}a, Grzegorz  and
      Alishahi, Afra",
    booktitle = "Proceedings of the 2018 {EMNLP} Workshop {B}lackbox{NLP}: Analyzing and Interpreting Neural Networks for {NLP}",
    month = nov,
    year = "2018",
    address = "Brussels, Belgium",
    publisher = "Association for Computational Linguistics",
    url = "https://aclanthology.org/W18-5446/",
    doi = "10.18653/v1/W18-5446",
    pages = "353--355"
}

@book{cover1999elements,
  title={Elements of Information Theory},
  author={Cover, Thomas M and Thomas, Joy A},
  year={1999},
  publisher={Wiley-Interscience}
}

@inproceedings{chen2020simple,
  title={A simple framework for contrastive learning of visual representations},
  author={Chen, Ting and Kornblith, Simon and Norouzi, Mohammad and Hinton, Geoffrey},
  booktitle={International conference on machine learning},
  pages={1597--1607},
  year={2020},
  organization={PmLR}
}

@inproceedings{he2020momentum,
  title={Momentum contrast for unsupervised visual representation learning},
  author={He, Kaiming and Fan, Haoqi and Wu, Yuxin and Xie, Saining and Girshick, Ross},
  booktitle={Proceedings of the IEEE/CVF conference on computer vision and pattern recognition},
  pages={9729--9738},
  year={2020}
}

@inproceedings{sun2016deep,
  title={Deep coral: Correlation alignment for deep domain adaptation},
  author={Sun, Baochen and Saenko, Kate},
  booktitle={European conference on computer vision},
  pages={443--450},
  year={2016},
  organization={Springer}
}
\bibliographystyle{iclr2026_conference}

\appendix
\newpage
\appendix
\onecolumn
\section{LLM Usage Statement}
We used LLMs solely to assist in drafting and polishing the writing of this
paper, without any other purposes.

\section{Proof of Theorem~1}
\label{app:proof_theorem1}

\noindent\textbf{Notation Setup.} Let $K=[k_1,\dots,k_N]\in\mathbb{R}^{d\times
N}$ and $R=[r_1,\dots,r_N]\in\mathbb{R}^{d\times N}$, and define
\begin{equation}
M := \sum_{i=1}^N k_i k_i^\top = K K^\top, \qquad
N := \sum_{i=1}^N r_i k_i^\top = R K^\top.
\end{equation}
Assume that $M$ is (pseudo-)invertible and define the closed-form solution
$\Delta = N M^+ = R K^\top (K K^\top)^+$, so that $\Delta \sum_i k_i k_i^\top =
\sum_i r_i k_i^\top$.

\noindent\textbf{Derivation of the Column Expansion $\Delta k_i = \sum_j
\beta_{ij} r_j$.} By definition,
\begin{equation}
\Delta = \sum_{j=1}^N r_j k_j^\top M^+ = R K^\top M^+.
\end{equation}
Applying $\Delta$ to a column $k_i$ gives
\begin{equation}
\Delta k_i = \sum_{j=1}^N r_j k_j^\top M^+ k_i.
\end{equation}
Setting $\beta_{ij} := k_j^\top M^+ k_i$, we immediately obtain
\begin{equation}
\Delta k_i = \sum_{j=1}^N \beta_{ij} r_j.
\end{equation}
This formula provides an explicit linear combination of residual embeddings
$r_j$ that reconstructs $\Delta k_i$, with coefficients $\beta_{ij}$ determined
by the key embeddings and the pseudo-inverse of $M$.

\noindent\textbf{Reconstruction Residual and Neighborhood Decomposition.} For
each knowledge item $i$, define the reconstruction residual $\mathrm{e}_i :=
\Delta k_i - r_i$. Suppose that $r_i$ can be approximately reconstructed from
its neighbors with nonnegative weights $q_{ij}$ for $j\ne i$, i.e.,
\begin{equation}
r_i = \sum_{j\ne i} q_{ij}\, r_j + \varepsilon_i, \qquad q_{ij}\ge0, \; \sum_{j\ne i} q_{ij}=1,
\end{equation}
where $\varepsilon_i$ denotes the residual error. Substituting this
decomposition into $\mathrm{e}_i$ gives
\begin{equation}
\mathrm{e}_i = \sum_{j\ne i} (\beta_{ij}-q_{ij})\, r_j + \beta_{ii} r_i - \varepsilon_i.
\end{equation}

\noindent\textbf{Bounding the Reconstruction Residual.} Taking norms and
applying the triangle inequality yields
\begin{equation}
\|\mathrm{e}_i\| \le \sum_{j\ne i} |\beta_{ij}-q_{ij}|\, \|r_j\| + |\beta_{ii}|\,\|r_i\| + \|\varepsilon_i\|.
\end{equation}
To relate the first term to embedding alignment, we construct a probability
vector $p_i$ from the positive parts of the coefficients $\beta_{ij}$ (for $j\ne
i$):
\begin{equation}
s_{ij} := \max\{\beta_{ij},0\}, \qquad S_i := \sum_{j\ne i} s_{ij}, \qquad p_{ij} := \frac{s_{ij}}{S_i}.
\end{equation}
Defining $C_i := \sum_{j\ne i} \|r_j\|$, one can show that
\begin{equation}
\sum_{j\ne i} |\beta_{ij}-q_{ij}|\,\|r_j\| \le C_i\, \mathrm{TV}(p_i, q_i),
\end{equation}
up to negligible contributions from negative $\beta_{ij}$ that can be absorbed
into $C_i$. Here, $\mathrm{TV}(p_i,q_i)$ is the \emph{total variation (TV)
distance} between two discrete distributions $p_i$ and $q_i$:
\begin{equation}
\mathrm{TV}(p,q) := \tfrac{1}{2}\sum_j |p_j - q_j|.
\end{equation}

Finally, applying Pinsker's inequality $\mathrm{TV}(p_i,q_i) \le
\sqrt{\tfrac12\, \mathrm{KL}(q_i\|p_i)}$ \citep{cover1999elements} gives
\begin{equation}
\|\mathrm{e}_i\| \le C_i\, \sqrt{\frac{1}{2}\, \mathrm{KL}(q_i\|p_i)} + |\beta_{ii}|\,\|r_i\| + \|\varepsilon_i\|.
\end{equation}
Identifying $q_i = P_r^{(i)}$ and $p_i = P_k^{(i)}$ with the kernel-normalized
neighborhood distributions from Definition~1 yields the embedding-alignment
bound stated in the main text:
\begin{equation}
\|\mathrm{e}_i\| \le C_i \sqrt{\frac12\, \mathcal{A}(i)} + |\beta_{ii}|\,\|r_i\| + \|\varepsilon_i\|.
\end{equation}

If $M$ is singular, replace $M^+$ with the Moore--Penrose pseudoinverse. The
contributions from negative $\beta_{ij}$ or scaling factors can usually be
absorbed into $C_i$. In the ideal case of perfect neighborhood alignment
$\mathcal{A}(i)=0$, negligible self-weight $\beta_{ii}=0$, and vanishing
residual $\varepsilon_i=0$, we recover $\mathrm{e}_i = 0$.

\section{Detailed Analysis of Corollary 2}
\label{app:same-subject-analysis}
We provide a detailed theoretical analysis that develops Corollary 2.

\textbf{Residual Decomposition.} Recall that the reconstruction residual can be
written as
\begin{equation}
e_i = \Delta k_i - r_i
    = \sum_{j=1}^N \beta_{ij} r_j - r_i,
\qquad
\beta_{ij} := k_j^\top\!\Big(\sum_{\ell=1}^N k_\ell k_\ell^\top\Big)^{\!-1}\! k_i.
\end{equation}
Partition the index set into the subject cluster $\mathcal S$ (samples sharing
the subject with $i$) and the remainder $\mathcal T$. Then
\begin{equation}\label{eq:resid-samesubject}
\Delta k_i
= \beta_{ii} r_i + \sum_{j \in \mathcal S, j \neq i} \beta_{ij} r_j
  + \sum_{j \in \mathcal T} \beta_{ij} r_j.
\end{equation}
Empirically and theoretically, only coefficients $\beta_{ij}$ for $j\in\mathcal
S$ become significant, while cross-subject coefficients remain negligible since
embeddings from different subjects are nearly orthogonal in key space and thus
contribute little to the reconstruction. Hence reconstruction is dominated by
the same-subject neighborhood.

\textbf{Effect of Adding Same-Subject Samples.} Let $K=\sum_\ell k_\ell
k_\ell^\top$ denote the Gram matrix. Suppose we add one additional key $k_j$
(with $j\in\mathcal S, j\ne i$). By the Woodbury identity\citep{horn1985matrix},
\begin{equation}
(K + k_jk_j^\top)^{-1}
= K^{-1} - \frac{K^{-1}k_j k_j^\top K^{-1}}{1 + k_j^\top K^{-1}k_j}.
\end{equation}
Consequently, the updated self-weight becomes
\begin{equation}\label{eq:beta-update}
\beta_{ii}^{\text{new}}
= k_i^\top (K + k_jk_j^\top)^{-1} k_i
= \beta_{ii} - \frac{(k_i^\top K^{-1} k_j)^2}{1 + k_j^\top K^{-1} k_j}.
\end{equation}
Thus $\beta_{ii}$ monotonically decreases as more same-subject vectors are
included. The lost weight is redistributed into off-diagonal terms
$\beta_{ij}$, consistent with our empirical observation in
Figure~\ref{fig:multiple_subjects}.

\textbf{Connection to Alignment.} Using the decomposition in
\eqref{eq:resid-samesubject}, the residual can be expressed as
\begin{equation}
e_i = \sum_{j\in\mathcal S} \beta_{ij}(r_j-r_i)
      + \sum_{j\in\mathcal T}\beta_{ij} r_j.
\end{equation}
Applying the triangle inequality yields
\begin{equation}\label{eq:error-bound}
\|e_i\|
\le \sum_{j\in\mathcal S} |\beta_{ij}|\,\|r_j-r_i\|
   + \sum_{j\in\mathcal T} |\beta_{ij}|\,\|r_j\|.
\end{equation}
The first term depends on the dispersion of responses within the same subject.
This dispersion is controlled by the alignment measure $\mathcal A(i)$: when
$\mathcal A(i)$ is small, the responses $\{r_j:j\in\mathcal S\}$ are tightly
clustered around $r_i$, so even a redistribution of weight from $\beta_{ii}$ to
other $\beta_{ij}$ produces only minor error. Conversely, when $\mathcal A(i)$
is large, intra-subject responses differ substantially, and the redistributed
weights amplify reconstruction error.

Combining \eqref{eq:beta-update} and \eqref{eq:error-bound}, we conclude that
co-editing additional same-subject samples (i) monotonically decreases
$\beta_{ii}$, (ii) redistributes weight into off-diagonal $\beta_{ij}$, and
(iii) yields residuals bounded by the intra-subject alignment $\mathcal A(i)$.
Therefore, massive editing performance crucially depends on the degree of
alignment within the subject cluster.

\section{Detailed Experiment Setup}
\label{appendix:exp_setup}
In the following, we provide detailed experimental configurations, including the
description of the datasets, introduction of baselines, explanation of
evaluation metrics, and implementation details.

\subsection{Datasets and Baselines}
\label{appendix:datasets_baselines}
We evaluate the performance of model editing techniques using the following
datasets:

\begin{itemize}[leftmargin=1em]
    \item \textbf{CounterFact}~\citep{meng2022locating} is a benchmark for
    evaluating factual knowledge localization and editing in LLMs. It contains
    21{,}917 entries that describe the named entities along with their
    counterfactual variations. Model editing techniques could be evaluated in
    terms of editing efficacy, generalization, and locality. The benchmark also
    contains generation prompts to test the model's generation ability after
    editing.

    \item \textbf{ZsRE}~\citep{levy2017zero} is a question-answering (QA)
    benchmark designed to evaluate zero-shot relation extraction capabilities of
    language models. Entries in the benchmark consist of a subject entity along
    with an answer as the editing target. The benchmark also includes
    paraphrased questions for testing generalization ability and irrelevant
    questions for evaluating the locality of editing techniques.

    \item \textbf{Wiki-recent}~\citep{zhang2024comprehensive} contains 1{,}266
    entries of triplets that have been added into \textsc{WikiData} after July
    2022. The benchmark enables insertion for models that were trained prior to
    the introduction of these facts. This simulates the cases of editing
    outdated models with newly introduced facts. Model editing techniques are
    evaluated in terms of editing efficacy, portability, and locality. Here,
    portability emphasizes whether the edited model could reason about the
    downstream effects of facts when they are inserted into the model.
\end{itemize}

We proceed to introduce baseline methods evaluated in the paper. For all
baseline methods, we use the official implementation provided by the authors.

\begin{itemize}[leftmargin=1em]
    \item \textbf{MEND}~\citep{mitchell2022fast} requires extra parameters for
    efficiently editing pretrained LLMs. It introduces a set of small auxiliary
    networks that transform standard fine-tuning gradients into low-rank
    updates, enabling fast and localized edits without retraining the entire
    model. This approach offers a scalable solution for post-hoc model editing,
    avoiding the overfitting issue of traditional fine-tuning methods.

    \item \textbf{ROME}~\citep{meng2022locating} performs factual knowledge
    editing by directly modifying the feed-forward weights in specific layers of
    LLMs. It first identifies that factual knowledge is primarily stored in
    mid-layer feed-forward modules, thereby demonstrating the feasibility of
    editing model parameters to update internal knowledge. The method then
    updates these weights to encode specific factual associations. ROME achieves
    precise insertion of new facts with minimal interference to unrelated
    knowledge. When evaluating using ROME, we edit all facts sequentially with
    batch size 1, as it does not support batch editing.

    \item \textbf{MEMIT}~\citep{meng2023memit} is designed to efficiently update LLMs with thousands of
    factual associations simultaneously. Building upon ROME, MEMIT employs a
    least-squares optimization over multiple key-value memory components, ensuring
    high specificity and minimal interference with unrelated knowledge. It further
    distributes the updates across multiple layers, which helps reduce the impact on
    the model's general capabilities.

    \item \textbf{PMET}~\citep{10.1609/aaai.v38i17.29818} is a method designed to
    enhance the precision of knowledge updates in large language models. Unlike
    prior approaches that treat transformer layer (TL) hidden states as direct
    inputs of the feed-forward network (FFN), PMET recognizes that these hidden
    states also encompass information from multi-head self-attention (MHSA) and
    residual connections. PMET proceeds to simultaneously optimize MHSA and FFN
    hidden states and use the optimized TC hidden states of FFN to precisely
    update FFN weights. This approach enables more accurate and efficient model
    editing, preserving the integrity of the model's existing knowledge while
    incorporating new information.

    \item \textbf{ALPHAEDIT}~\citep{fang2024alphaedit} preserves knowledge in LLMs during sequential updates
    by projecting updates onto the null space of the preserved knowledge. This could
    ensure that new modifications do not interfere with previously stored
    information. This approach maintains the integrity of the model's existing
    knowledge while enabling precise edits.
\end{itemize}

\subsection{Metrics}
\label{appendix:metrics}
We now introduce the metrics used for CounterFact, Wiki-recent and ZsRE
respectively.

\subsubsection{CounterFact Metrics}
Given an LLM $f_\theta$, a knowledge fact tuple (subject $s_i$, relation $r_i)$,
a target output $o_i$ and the original output $o_i^c$, we define the following
metrics:
\begin{itemize}[leftmargin=1em]
    \item \textbf{Editing Efficacy}: Unlike previous works that evaluate the
    portion of cases where $o_i$ is more probable than $o_i^c$, we directly
    compute the average top-1 accuracy of edited samples.
    \begin{equation}
        \mathds{E}_i[o_i=\arg\mathop{\max}_{o} \mathds{P}_{f_\theta}(o\ |\ (s_i,r_i))]
    \end{equation}

    \item \textbf{Generalization}: Average top-1 accuracy of the edited model on rephrased
    statements $N((s_i,r_i))$ of the original knowledge fact. Rephrased statements
    share the same semantic meaning with the original statements.
    \begin{equation}
        \mathds{E}_i[o_i=\arg\mathop{\max}_{o} \mathds{P}_{f_\theta}(o\ |\ N((s_i,r_i)))]
    \end{equation}

    \item \textbf{Specificity}: The portion of cases where $o_i^c$ is more probable than
    $o_i$ with neighboring statements $O((s_i,r_i))$. Neighboring statements are
    constructed using prompts which share distinct but semantically related subjects
    with the original knowledge fact.
    \begin{equation}
        \mathds{E}_i[\mathds{P}_{f_\theta}(o_i^c\ |\ O((s_i,r_i))) > \mathds{P}_{f_\theta}(o_i\ |\ O((s_i,r_i)))]
    \end{equation}

    \item \textbf{Fluency}: Fluency score measures the quality of the generated text. It
    scores low if the generated text contains excessive repetition.
    \begin{equation}
        -\frac{2}{3}\mathop{\sum}_{k}g_2(k)\log_2g_2(k)+\frac{4}{3}\mathop{\sum}_{k}g_3(k)\log_2g_3(k)
    \end{equation}
    where $g_2(k)$ and $g_3(k)$ are the probabilities of bigram and trigram $k$
    respectively.
\end{itemize}

\subsubsection{Wiki-recent Metrics}
Given a LLM $f_\theta$, a knowledge fact tuple $(s_i,\ r_i)$, a target output
$o_i$ and the original output $o_i^c$, we define the following metrics:

\begin{itemize}[leftmargin=1em]
    \item \textbf{Editing Efficacy}: Average top-1 accuracy of edited samples.
    \begin{equation}
        \mathds{E}_i[o_i=\arg\mathop{\max}_{o} \mathds{P}_{f_\theta}(o\ |\ (s_i,r_i))]
    \end{equation}

    \item \textbf{Portability}: Average top-1 accuracy of the edited model on
    portability prompts $P((s_i,r_i))$ of the original knowledge fact.
    Portability prompts contain three parts: alias prompts, compositionality
    and reasoning prompts, and logical generation prompts. Specifically, alias
    prompts are constructed by replacing the subject $s_i$ with an alias or
    synonym. Compositionality and reasoning prompts require the post-edit model
    to conduct reasoning about the changed fact. Logical generation prompts are
    changes that are semantically related to the modified fact and expected to
    change by the edit.

    \begin{equation}
        \mathds{E}_i[o_i=\arg\mathop{\max}_{o} \mathds{P}_{f_\theta}(o\ |\ P((s_i,r_i)))]
    \end{equation}

    \item \textbf{Locality}: Average top-1 accuracy of the edited model on neighboring prompts
    $O((s_i,r_i))$ of the original knowledge fact.
    \begin{equation}
        \mathds{E}_i[o_i=\arg\mathop{\max}_{o} \mathds{P}_{f_\theta}(o\ |\ O((s_i,r_i)))]
    \end{equation}

    \item \textbf{Fluency}: Fluency score measures the quality of the generated text. It
    scores low if the generated text contains excessive repetition.
    \begin{equation}
        -\frac{2}{3}\mathop{\sum}_{k}g_2(k)\log_2g_2(k)+\frac{4}{3}\mathop{\sum}_{k}g_3(k)\log_2g_3(k)
    \end{equation}
    where $g_2(k)$ and $g_3(k)$ are the probabilities of bigram and trigram $k$
    respectively.
\end{itemize}

\subsubsection{ZsRE Metrics}
Given a LLM $f_\theta$, a knowledge fact tuple $(s_i,\ r_i)$, a target output
$o_i$ and the original output $o_i^c$, we define the following metrics:

\begin{itemize}[leftmargin=1em]
    \item \textbf{Editing Efficacy}: Average top-1 accuracy of edited samples.
    \begin{equation}
        \mathds{E}_i[o_i=\arg\mathop{\max}_{o} \mathds{P}_{f_\theta}(o\ |\ (s_i,r_i))]
    \end{equation}

    \item \textbf{Generalization}: Average top-1 accuracy of the edited model on
    generalization prompts $N((s_i,r_i))$ of the original knowledge fact.
    \begin{equation}
        \mathds{E}_i[o_i=\arg\mathop{\max}_{o} \mathds{P}_{f_\theta}(o\ |\ N((s_i,r_i)))]
    \end{equation}

    \item \textbf{Locality}: Average top-1 accuracy of the edited model on neighboring prompts
    $O((s_i,r_i))$ of the original knowledge fact.
    \begin{equation}
        \mathds{E}_i[o_i=\arg\mathop{\max}_{o} \mathds{P}_{f_\theta}(o\ |\ O((s_i,r_i)))]
    \end{equation}
\end{itemize}

\subsubsection{Examples of Evaluation Prompts with Prefixes}
\label{appendix:examples_of_prompts}
We provide examples of evaluation prompts with prefixes in the following. For
the evaluated prompt "The mother tongue of Danielle Darrieux is French",
corresponding evaluated prompts with ten distinct 5-token prefixes are:

\begin{tcolorbox}[
    colback=gray!5,
    colframe=gray!50,
    arc=0mm,
    title=Example Prompts with Prefixes,
    fonttitle=\bfseries
]
\begin{itemize}[leftmargin=1em]
    \item The doctor's office was too. The mother tongue of Danielle Darrieux is French.
    \item Therefore, the average speed of. The mother tongue of Danielle Darrieux is French.
    \item You can find many examples of. The mother tongue of Danielle Darrieux is French.
    \item However, the government's new. The mother tongue of Danielle Darrieux is French.
    \item And so, as the night. The mother tongue of Danielle Darrieux is French.
    \item While the world of sports and. The mother tongue of Danielle Darrieux is French.
    \item To make a cake, you. The mother tongue of Danielle Darrieux is French.
    \item Nevertheless, the overall sentiment of. The mother tongue of Danielle Darrieux is French.
    \item Never tried it before. The mother tongue of Danielle Darrieux is French.
    \item He realized that the people of. The mother tongue of Danielle Darrieux is French.
\end{itemize}
\end{tcolorbox}
For model evaluation, we generate prefixes by prompting the unedited model with
the following initial words: "The", "Therefore", "You", "However", "And",
"While", "To", "Nevertheless", "Never", and "He". These initial words are used
to generate diverse 5-token prefixes, which are then prepended to each edited
fact during the evaluation process. This approach ensures a comprehensive
assessment of the model's performance across different linguistic contexts.

\subsection{Implementation Details}
\label{implementation_details}
We implement all experiments on a single NVIDIA H800 GPU with 80GB memory.
During optimization, we iterate for 25 steps with 0.5 learning rate. We set
$M=50$ for balancing the fine-grained alignment and optimization efficiency. The
details of our implementation across different models are outlined as follows:

\begin{itemize}[leftmargin=1em]
    \item \textbf{LLAMA2-7B}: We modify layers [3, 4] for editing factual knowledge. The
    hyperparameters $\lambda$s are set to [4000, 4000] respectively for two
    layers. We set $\lambda_{KL}=2$ and $\lambda_{MSE}=8$.

    \item \textbf{Qwen-2.5-7B}: We modify layers [3, 4] for editing factual knowledge. The
    hyperparameters $\lambda$s are set to [500, 500] respectively for two
    layers. We set $\lambda_{KL}=1.5$ and $\lambda_{MSE}=8$.

    \item \textbf{LLaMA2-13B}: We modify layers [3, 4] for editing factual knowledge. The
    hyperparameters $\lambda$s are set to [4000, 4000] respectively for two
    layers. We set $\lambda_{KL}=1.5$ and $\lambda_{MSE}=8$.

    \item \textbf{Falcon-7B}: We modify layers [3, 4] for editing factual knowledge. The
    hyperparameters $\lambda$s are set to [1000, 1000] respectively for two
    layers. We set $\lambda_{KL}=2$ and $\lambda_{MSE}=8$.

    \item \textbf{Deepseek-base-7B}: We modify layers [3, 4] for editing factual knowledge. The
    hyperparameters $\lambda$s are set to [4000, 4000] respectively for two
    layers. We set $\lambda_{KL}=4.5$ and $\lambda_{MSE}=8$.

    \item \textbf{LLaMA3-8B}: We modify layers [3, 4] for editing factual knowledge. The
    hyperparameters $\lambda$s are set to [1000, 1000] respectively for two
    layers. We set $\lambda_{KL}=2$ and $\lambda_{MSE}=8$.
\end{itemize}

We justify our choice of editing layers by comparing against the common settings
used in prior work
\citep{meng2023memit,10.1609/aaai.v38i17.29818,meng2022locating}. On LLaMA2-7B,
we evaluate EAMET with different layer combinations when editing 10,000 facts
from CounterFact and ZsRE. As shown in \cref{app:exp_massive_editing_layers},
EAMET achieves higher efficacy and generalization with layers [3, 4] compared to
[4, 5, 6, 7, 8].

\begin{table}[h!]
    \centering
    \caption{Performance comparison of EAMET on LLAMA2-7B with different layer selections.} 
    \scalebox{1}{
    \begin{tabular}{c|cccc|ccc}
    \toprule
    \multicolumn{1}{c}{\multirow{2}{*}{\textbf{Layers}}} &
    \multicolumn{4}{c}{\textbf{Counterfact}} &
    \multicolumn{3}{c}{\textbf{ZsRE}} \\
    \cmidrule(lr){2-5}\cmidrule(lr){6-8}
    \multicolumn{1}{c}{} & \textbf{Eff.}$\uparrow$ & \textbf{Gen.}$\uparrow$ & \textbf{Spe.}$\uparrow$ & \textbf{Flu.}$\uparrow$ & \textbf{Eff.}$\uparrow$ & \textbf{Gen.}$\uparrow$ & \textbf{Spe.}$\uparrow$ \\
    \midrule
    3, 4 & 89.09 & 61.21 & 72.19 & 519.23 & 89.47 & 81.34 & 15.70 \\
    4, 5, 6, 7, 8 & 77.58 & 36.83 & 73.43 & 516.63 & 87.14 & 76.91 & 15.92 \\
    \bottomrule
    \end{tabular}
    }
    \label{app:exp_massive_editing_layers}
\end{table}

\begin{figure}[!h]
    \begin{center}
    \scalebox{1}{
    \centerline{\includegraphics[width=\columnwidth]{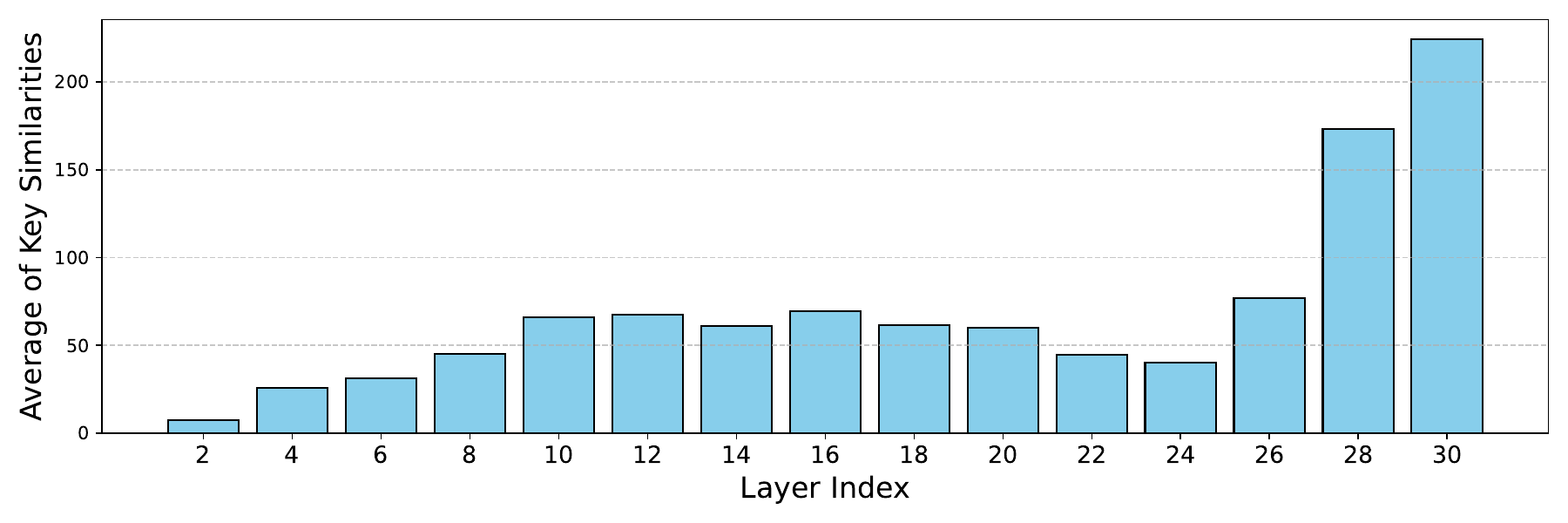}}
    }
    \caption{Average of key similarities across different layers of LLaMA2-7B when editing 500 knowledge items.}
    \label{app:ks_similarities_by_layer}
    \end{center}
\end{figure}

This effect arises because, as the edited layer becomes deeper, the similarity
between key embeddings of different knowledge items increases. As shown in
\cref{app:ks_similarities_by_layer}, the average similarity across layers of
LLaMA2-7B grows with layer depth. When the last edited layer is $8$, the average
similarity is nearly twice that of layer $4$. This growth makes it more
difficult to align the memory embedding space with the key embedding space,
since KL divergence primarily captures distributional differences and we only
apply MSE loss to the top-$M$ cosine similarities. Applying MSE to all cosine
similarities may lead to vanishing gradients. As the number of similarities
grows, the strongest ones become diluted, which slows convergence and hinders
optimization. Such misaligned memory embeddings can substantially degrade both
the effectiveness and robustness of massive editing.

\section{Algorithmic Description of EAMET}
\label{appendix:full_algorithm}

\begin{algorithm}[h!]
    \caption{The EAMET Algorithm}
    \label{alg:EAMET}
    \DontPrintSemicolon
    \SetKwFor{For}{for}{do}{end}

    \textbf{Data}: Requested edits $\mathcal{E} = \{(s_i, rel_i, o_i)\}$, generator $G$, layers to edit $\mathcal{S}$, covariances $C^l$\;
    \textbf{Result}: Modified generator containing edits from $\mathcal{E}$\;
    \;
    $K \leftarrow []$ \;
    $L \leftarrow \text{final layer of candidate layers }\mathcal{R}$ \;
    \For{$s_i, rel_i, o_i \in \mathcal{E}$}{
        $k_i^L \leftarrow k_i^L = \frac{1}{N_{FP}} \sum_{j=1}^{N_{FP}} k(f_j \oplus s_i)$ \;
        $K \leftarrow K \cup \{k_i^l\}$ \;
    }
    $P_k \leftarrow \{P_k^{(i,j)} = \cos(k_i, k_j) \mid j \ne i, k_i, k_j\in K\}$ \;
    $R \leftarrow []$ \;
    \For{$s_i, rel_i, o_i \in \mathcal{E}$}{
        $r_i \leftarrow h_i^L$ \tcp*[r]{Initialize $r_i$ as the original hidden state}
        $P_r^{(i)} \leftarrow \{P_r^{(i,j)} \mid j < i, r_j\in R\}$ \;
        $\bar{P}_k^{(i)} \leftarrow \{P_k^{(i,j)} \mid j < i, k_j\in K\}$ \;
        $L_{\text{KL}}(i) = \text{KL}\!\left(P_r^{(i)} \,\|\,\bar{P}_k^{(i)}\right)$ \;
        $I_K \leftarrow \text{indices of the top } M \text{ largest elements in } \bar{P}_k^{(i)}$ \;
        $L_{\text{MSE}}(i) = \frac{1}{M} \sum_{j\in I_K} \big\| P_r^{(i,j)} - P_k^{(i,j)} \big\|^2$ \;
        $r_i \leftarrow \arg\min_{r_i} \frac{1}{N_{FP}} \sum_{j=1}^{N_{FP}} -\log \mathbb{P}_{G_{(h_i^L+=r_i)}}[o_i \mid f_j \oplus tp(s_i, r_i)]+\lambda_{KL}L_{\text{KL}}(i)+\lambda_{MSE}L_{\text{MSE}}(i)$ \;
        $z_i \leftarrow h_i^L + r_i$ \;
        $R \leftarrow R \cup \{r_i\}$ \;
    }

    \For{$l \in \mathcal{R}$}{
        $h_i^l \leftarrow h_i^{l-1} + a_i^l + m_i^l$ \;
        \For{$s_i, rel_i, o_i \in \mathcal{E}$}{
            $k_i^l \leftarrow k_i^l = \frac{1}{N_{FP}} \sum_{j=1}^{N_{FP}} k(f_j \oplus s_i)$ \;
            $r_i^l \leftarrow \frac{z_i-h_i^L}{L-l+1}$ \tcp*[r]{Distribute over remaining layers}
        }
        $K^l \leftarrow [k_1^l, \ldots, k_{N_t}^l]$ \;
        $R^l \leftarrow [r_1^l, \ldots, r_{N_t}^l]$ \;
        $\Delta \leftarrow R^l K^{l^T}(C^l + K^l K^{l^T})^{-1}$ \;
        $W^l \leftarrow W^l + \Delta$ \tcp*[r]{Update layer $l$ MLP weights in model}
    }
\end{algorithm}
In this section, we present a detailed description of the EAMET algorithm in
Algorithm~\ref{alg:EAMET}. The procedure consists of three main stages: (1) key
embedding preparation, (2) aligning memory embeddings with key embeddings, and
(3) distributing MLP updates across candidate layers.

\noindent \textbf{Key Embedding Preparation}. Following prior work
\citep{meng2022locating, meng2023memit,10.1609/aaai.v38i17.29818}, we evenly
distribute MLP updates across the critical target layers $\mathcal{R}$. We
denote by $L$ the final candidate layer where new memories are fully represented
(Line 5). Before optimizing residual embeddings, we first compute the key
embeddings for each target edit (Line 7). To improve generalization, each
subject is augmented with $N_{FP}-1$ random prefixes of fixed length $f_i$. All
resulting key embeddings are aggregated into a matrix $K$ (Line 8). We then
compute pairwise cosine similarities among key embeddings to obtain $P_k$ (Line
9).

\noindent \textbf{Aligning Memory Embeddings with Key Embeddings}. For each
target edit, we initialize the residual embedding with the original hidden state
(Line 12), since it naturally corresponds to the associated key embedding and
thus provides a good starting point. We compute cosine similarities among
residual embeddings to form $P_r^{(i)}$ (Line 13), and extract the corresponding
key embedding structure $\bar{P}_k^{(i)}$ (Line 14). Alignment is achieved by
minimizing the KL divergence between $P_r^{(i)}$ and $\bar{P}_k^{(i)}$ (Line
15). To further refine alignment, we select the indices $I_K$ corresponding to
the top $M$ similarities in $P_k^{(i)}$ (Line 16) and minimize the MSE loss
between $P_r^{(i)}$ and $P_k^{(i)}$ restricted to these indices (Line 17). The
optimized residual embedding $r_i$ is obtained by minimizing this combined
objective (Line 18) and stored in the set $R$ (Line 19).

\noindent \textbf{Distributing MLP Updates Across Candidate Layers}. We update
MLP modules sequentially across layers $l \in \mathcal{R}$, as earlier edits
affect subsequent representations. For each candidate layer, we compute key
embeddings $k_i^l$ for all edits (Line 24) and residual embeddings $r_i^l$,
distributing them proportionally across layers (Line 25). These embeddings are
then aggregated into $K^l$ and $R^l$ (Lines 26-27) to update the MLP weights
with $\Delta$ (Lines 28-29).

\section{Additional Experimental Results}
\label{appendix:additional_results}
In this section, we present additional experiments and findings to further
validate the effectiveness of EAMET. We begin by evaluating editing performance
across different semantic categories. Next, we assess its impact on the model's
general capabilities using the GLUE benchmarks. We then report results on two
additional LLMs, Gemma-7B \citep{team2024gemma} and Phi-1.5
\citep{li2023textbooks}. We also examine how the order of edits affects EAMET's
performance. Furthermore, we explore its integration with sequential editing,
showing that embedding alignment enables larger batch sizes per step and thus
reduces the number of steps needed to edit the same set of knowledge items.
Finally, we provide an ablation study on combining KL loss and MSE loss, along
with a comprehensive analysis of the hyperparameters $\lambda_{KL}$,
$\lambda_{MSE}$, and $M$.

\subsection{Editing Performance Involving Different Semantics}
\label{appendix:semantics}
\begin{figure}[!h]
    \begin{center}
    \scalebox{1}{
    \centerline{\includegraphics[width=\columnwidth]{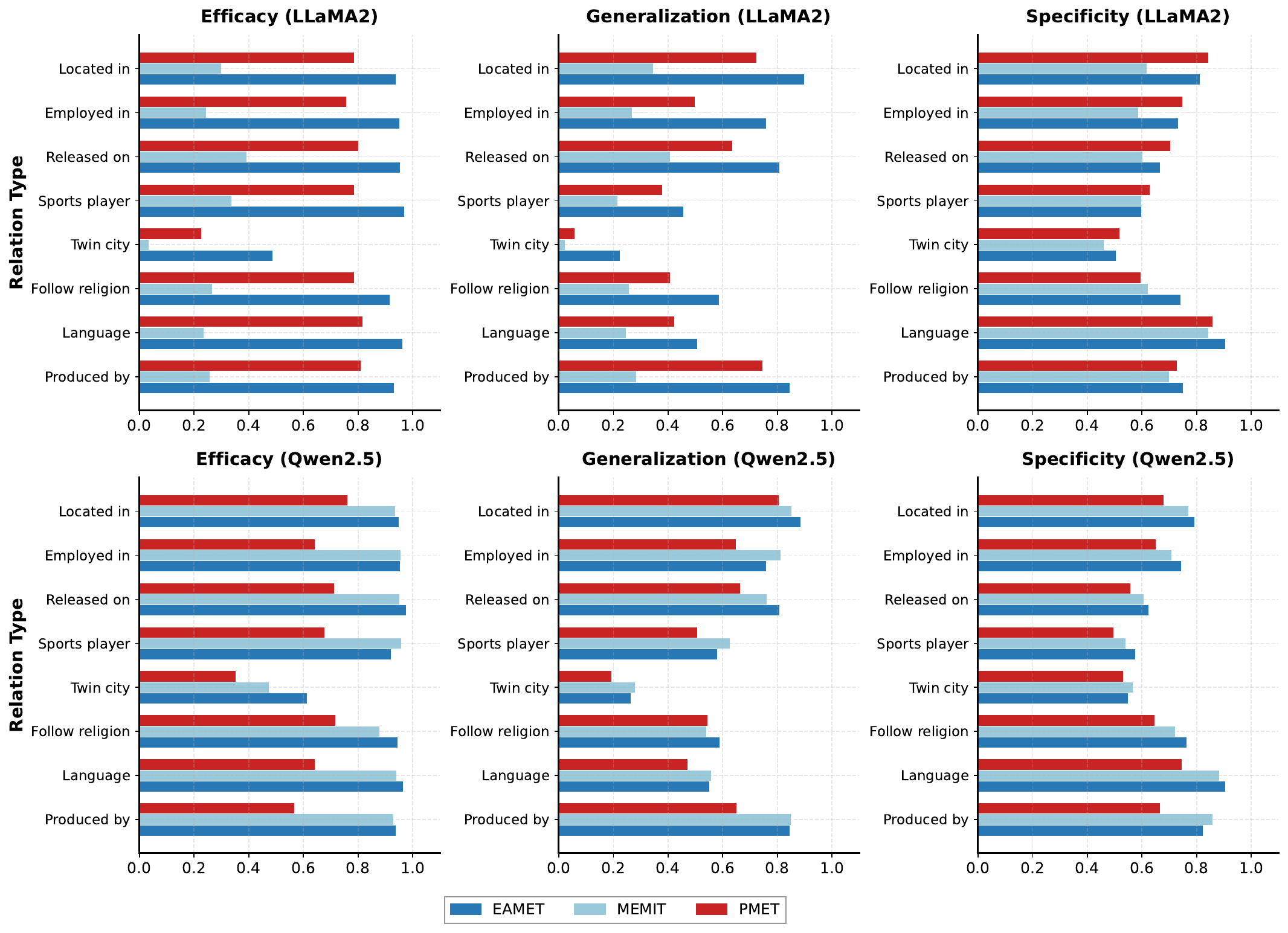}}
    }
    \caption{Performance comparison of different editing methods across different semantics.}
    \label{fig:model_comparison_metrics}
    \end{center}
\end{figure}

\textbf{\textit{Additional Finding 1.} EAMET Achieves Superior Editing
Performance Across Different Semantics.} We extract samples with specific
relation types from the CounterFact dataset to evaluate the performance of
different editing methods across semantic categories. As shown in
\cref{fig:model_comparison_metrics}, EAMET consistently achieves the highest
editing efficacy and generalization on both LLaMA2-7B and Qwen2.5 for most
semantic types. On LLaMA2-7B, EAMET outperforms the second-best method (PMET) by
approximately 10\% in efficacy and 20\% in generalization. 
In terms of editing specificity, EAMET performs better on Qwen2.5 than on
LLaMA2. On LLaMA2, PMET surpasses EAMET by an average of 5\%, whereas on
Qwen2.5, EAMET achieves the highest specificity on 6 out of 8 relation types.

We observe an interesting phenomenon on the \textit{twin-city} relation of LLaMA2-7B:
EAMET achieves 2$\times$ higher efficacy and 4$\times$ higher
generalization compared to PMET, while MEMIT nearly fails on this relation,
yielding efficacy and generalization scores close to 0\%. This occurs because
facts involving the twin-city relation are typically expressed in forms such as
“The twin city of {subject}” or “What is the twin city of {subject}?”. The key
embeddings, which are extracted from the last subject token, are therefore
highly similar across facts due to the shared prefixes in these templates. As a
result, reconstructing each individual update $\Delta k_i = r_i$ from the global
update $\Delta$ computed in \cref{final_MLP_update} requires proper alignment between
key embeddings and residual embeddings. Methods lacking this alignment
constraint struggle to separate the highly overlapping keys, leading to poor
performance on this relation.

\subsection{General Ability of Edited Models on GLUE Benchmarks}
\label{appendix:glue_benchmarks}
\begin{figure}[h]
    \centering
    \scalebox{1}{
        \includegraphics[width=\textwidth]{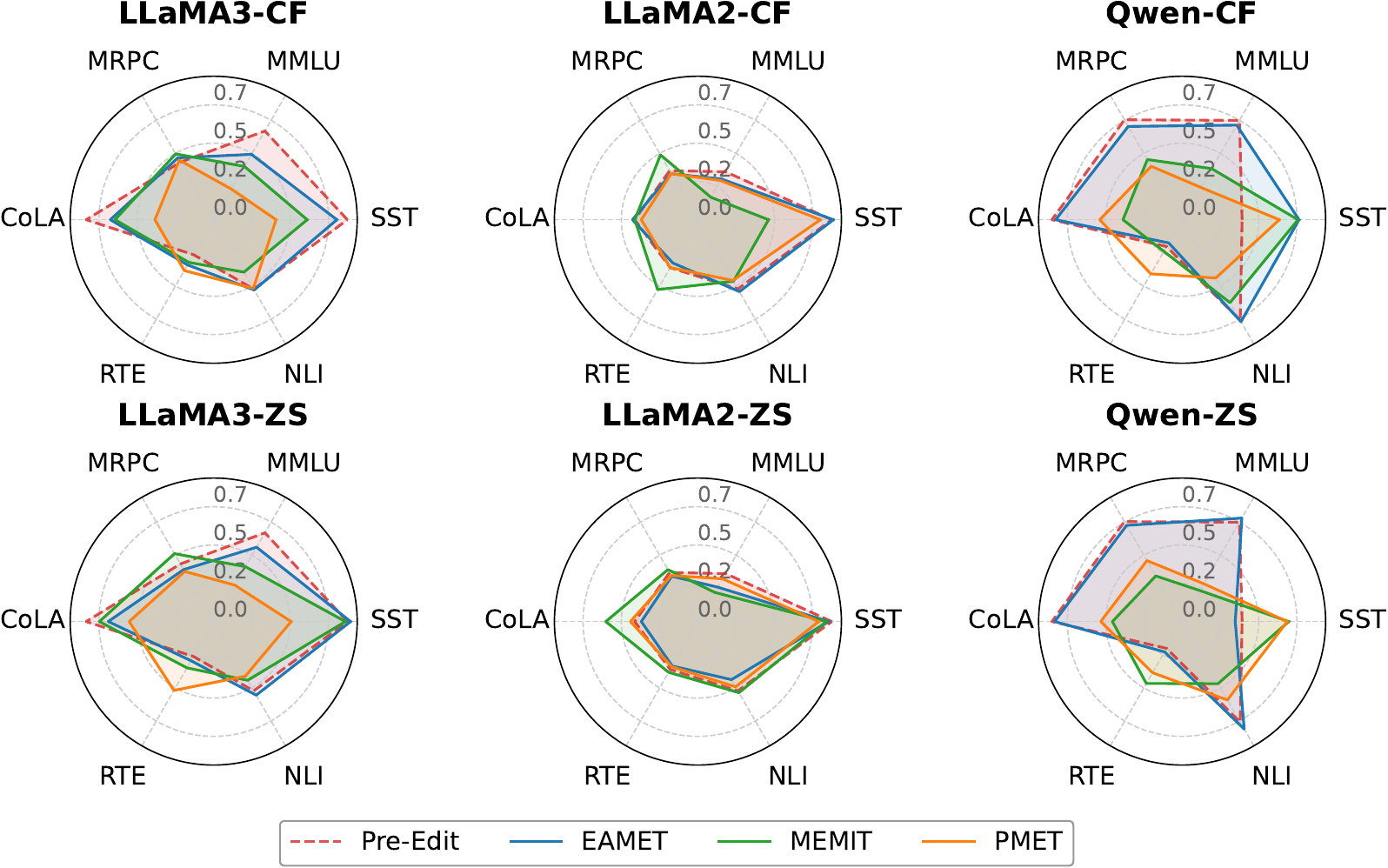}
    } \caption{General ability of pre-edited model and models edited by
    different methods on GLUE benchmarks}
    \label{fig:glue_benchmarks}
\end{figure}

\textbf{\textit{Additional Finding 2.} EAMET Better Preserves the General
Ability of LLMs After Massive Editing.} We examine whether large-scale editing
degrades the general capabilities of LLMs under MEMIT, PMET, and EAMET.
Specifically, we evaluate three models (LLaMA2-7B, Qwen2.5-7B, and LLaMA3-8B) on
the GLUE benchmark \citep{wang-etal-2018-glue} after editing 10,000 knowledge
facts from CounterFact and ZsRE. For reference, we also report the performance
of the unedited models. As shown in \cref{fig:glue_benchmarks}, EAMET
consistently yields the smallest performance deviation from pre-edit baselines.
On Qwen2.5-7B, the average deviation across six GLUE tasks is only 0.083 for
CounterFact and 0.032 for ZsRE, substantially lower than MEMIT (0.266 and 0.349)
and PMET (0.310 and 0.276). A similar trend holds for LLaMA3-8B and LLaMA2-7B:
on CounterFact, EAMET achieves average deviations of 0.083 and 0.025, compared
to 0.155 and 0.043 for MEMIT, the second-best method.

We attribute this robustness to \textsc{EAMET}'s ability to extract more aligned
memory representations across knowledge items. Such alignment reduces the
likelihood of embedding inconsistency between key and residual spaces during
massive editing, which may compromise the model's general capabilities. By
mitigating this interference, \textsc{EAMET} effectively preserves the original
functionality of the LLM.

\subsection{Evaluation on Additional LLMs}
\label{appendix:more_models}
\begin{table}[!h]
    \centering
    \caption{Performance comparison of different editing methods on Gemma-7B and Phi-1.5 on the Counterfact and ZsRE benchmarks.}
    \scalebox{1}{
    \begin{tabular}{cc|cccc|cccc|ccc}
    \toprule
    \multicolumn{1}{c}{\multirow{2}{*}{\textbf{Model}}} &
    \multicolumn{1}{c}{\multirow{2}{*}{\textbf{Method}}} &
    \multicolumn{4}{c}{\textbf{Counterfact}} &
    \multicolumn{3}{c}{\textbf{ZsRE}} \\
    \cmidrule(lr){3-6}\cmidrule(lr){7-9}
    \multicolumn{1}{c}{} & \multicolumn{1}{c}{} & \textbf{Eff.}$\uparrow$ & \textbf{Gen.}$\uparrow$ & \textbf{Spe.}$\uparrow$ & \textbf{Flu.}$\uparrow$ & \textbf{Eff.}$\uparrow$ & \textbf{Gen.}$\uparrow$ & \textbf{Spe.}$\uparrow$ \\
    \midrule
    \multirow{3}{*}{Gemma-7B} 
     & MEMIT & 93.01 & 54.33 & 74.88 & \textbf{538.92} & 84.61 & 73.66 & 23.04 \\
     & PMET & 91.69 & 47.24 & \textbf{75.22} & 533.62 & 83.05 & 73.95 & \textbf{23.85}\\
     & \textbf{EAMET} & \textbf{95.29} & \textbf{68.53} & 70.22 & 530.93 & \textbf{91.69} & \textbf{86.43} & 23.37\\
    \midrule
    \multirow{3}{*}{Phi-1.5}
     & MEMIT & 49.00 & 29.67 & 63.71 & 568.21 & 49.76 & 38.54 & 20.36 \\
     & PMET & 38.49 & 20.75 & \textbf{67.01} & 579.19 & 32.00 & 24.47 & \textbf{21.54} \\
     & \textbf{EAMET} & \textbf{67.76} & \textbf{40.13} & 62.08 & \textbf{580.92} & \textbf{60.61} & \textbf{45.42} & 21.29 \\
    \bottomrule
    \end{tabular}
    }
    \label{tab:more_models}
\end{table}

We further evaluate the performance of EAMET on two additional LLMs: Gemma-7B
and Phi-1.5. As shown in \cref{tab:more_models}, EAMET consistently achieves the
highest editing efficacy and generalization across both models and datasets.
Moreover, it maintains competitive performance in terms of editing locality and
generation ability.

\subsection{Integration with Sequential Editing}
\label{appendix:integration_sequential}

We further examine the impact of EAMET on sequential editing. We hypothesize
that incorporating embedding alignment can increase the effective batch size at
each step, thereby reducing the number of steps required to edit the same set of
knowledge items. To this end, we adopt the state-of-the-art sequential editing
method AlphaEdit, which preserves knowledge in LLMs by projecting updates onto
the null space of preserved knowledge. To evaluate the benefit of embedding
alignment, we replace AlphaEdit's target memory optimization with EAMET,
resulting in a variant we call AlphaEdit-Aligned. Importantly, this substitution
does not alter AlphaEdit's core design, since the method was not originally
tailored for optimizing target memory. We then compare AlphaEdit and
AlphaEdit-Aligned when editing 2,000 knowledge items on LLAMA2-7B from the
Counterfact and ZsRE datasets, varying the batch size across 100, 200, 400, and
500 to evaluate how batch size influences performance.

\begin{table}[!h]
    \centering
    \caption{Performance comparison of AlphaEdit and its version itegrated with
     EAMET on the Counterfact and ZsRE benchmarks.}
    \scalebox{0.9}{
    \begin{tabular}{cc|cccc|cccc|ccc}
    \toprule
    \multicolumn{1}{c}{\multirow{2}{*}{\textbf{Method}}} &
    \multicolumn{1}{c}{\multirow{2}{*}{\textbf{Batch Size}}} &
    \multicolumn{4}{c}{\textbf{Counterfact}} &
    \multicolumn{3}{c}{\textbf{ZsRE}} \\
    \cmidrule(lr){3-6}\cmidrule(lr){7-9}
    \multicolumn{1}{c}{} & \multicolumn{1}{c}{} & \textbf{Eff.}$\uparrow$ & \textbf{Gen.}$\uparrow$ & \textbf{Spe.}$\uparrow$ & \textbf{Flu.}$\uparrow$ & \textbf{Eff.}$\uparrow$ & \textbf{Gen.}$\uparrow$ & \textbf{Spe.}$\uparrow$ \\
    \midrule
    \multirow{4}{*}{AlphaEdit} 
     & 100 & 49.10 & 39.13 & 61.01 & 331.83 & 95.75 & 87.75 & 17.05 \\
     & 200 & 47.85 & 40.75 & 61.25 & 324.23 & 95.05 & 87.70 & 17.00\\
     & 400 & 41.55 & 37.03 & 59.51 & 228.53 & 94.80 & 86.75 & 16.80\\
     & 500 & 39.05 & 40.25 & 59.70 & 306.72 & 94.50 & 86.15 & 16.85\\
    \midrule
    \multirow{4}{*}{AlphaEdit-Aligned}
     & 100 & 96.75 & 66.13 & 66.48 & 505.59 & 96.55 & 87.75 & 17.00 \\
     & 200 & 96.45 & 65.73 & 66.44 & 505.47 & 96.80 & 87.30 & 16.95 \\
     & 400 & 96.45 & 64.85 & 66.33 & 505.82 & 95.61 & 86.95 & 16.80 \\
     & 500 & 96.40 & 65.66 & 66.38 & 506.63 & 95.50 & 87.15 & 16.85 \\
    \bottomrule
    \end{tabular}
    }
    \label{tab:integration_sequential}
\end{table}

\noindent \textbf{\textit{Additional Finding 3.} Integrating Embedding Alignment
with Sequential Editing Enables Larger Batch Sizes.} As shown in
\cref{tab:integration_sequential}, AlphaEdit-Aligned consistently outperforms
AlphaEdit across all batch sizes, indicating that embedding alignment
effectively enlarges the batch size per step. The improvement is especially
pronounced on the Counterfact dataset, where batch editing is notably more
difficult without aligning key and residual embeddings. These results suggest
that EAMET can be seamlessly integrated into sequential editing to further
enhance editing performance.

\subsection{Ablation Study}
\label{appendix:ablation}

\begin{table}[h!]
    \centering
    \caption{Ablation study of EAMET components on Counterfact and ZsRE datasets.}
    \scalebox{1}{
    \begin{tabular}{l|cccc|ccc}
    \toprule
    \multicolumn{1}{c}{\multirow{2}{*}{\textbf{Method}}} &
    \multicolumn{4}{c}{\textbf{Counterfact}} &
    \multicolumn{3}{c}{\textbf{ZsRE}} \\
    \cmidrule(lr){2-5}\cmidrule(lr){6-8}
    \multicolumn{1}{c}{} & \textbf{Eff.}$\uparrow$ & \textbf{Gen.}$\uparrow$ & \textbf{Spe.}$\uparrow$ & \textbf{Flu.}$\uparrow$ & \textbf{Eff.}$\uparrow$ & \textbf{Gen.}$\uparrow$ & \textbf{Spe.}$\uparrow$ \\
    \midrule
    EAMET (Full) & 89.09 & 61.21 & 72.19 & 519.42 & 89.47 & 81.34 & 15.70 \\
    w/o KL Loss & 83.45 & 60.16 & 71.70 & 519.78 & 88.16 & 80.46 & 15.40 \\
    w/o MSE Loss & 86.98 & 53.77 & 72.90 & 516.90 & 86.45 & 73.12 & 14.61 \\
    \bottomrule
    \end{tabular}
    }
    \label{tab:ablation_study}
\end{table}

We further justify the design of combining KL loss and MSE loss by conducting
ablations that remove either component. As shown in \cref{tab:ablation_study},
the full version of EAMET consistently achieves the best overall performance
across both datasets, while excluding either loss results in a clear performance
drop. This confirms the effectiveness of our joint loss design.

Interestingly, the two losses exhibit different levels of importance depending
on the dataset. On Counterfact, removing KL loss causes a 6\% drop in editing
efficacy, compared to only 2\% when removing MSE loss. In contrast, on ZsRE,
excluding KL loss leads to a minor 1\% drop, whereas removing MSE loss results
in a larger 3\% decline. This difference stems from the structure of the
datasets: in Counterfact, each knowledge item has a unique subject, making their
key embeddings nearly orthogonal (low cosine similarity). Here, KL loss, which
captures distributional differences across embeddings, plays a more critical
role, while MSE contributes less. In ZsRE, however, many items share the same
subject, leading to highly similar key embeddings (high cosine similarity). In
this case, MSE loss is more important, as it directly aligns residual embeddings
with their corresponding key embeddings within these subject-specific
neighborhoods.

\subsection{Efficiency Analysis}
\label{appendix:efficiency_analysis}

We provide anaysis on the practical deployment cost of EAMET compared with
MEMIT. We note that EAMET and MEMIT follow highly similar workflows for updating
knowledge in LLMs: both require per-fact residual optimization. EAMET involves
two additional steps: 1) the key embedding preparation stage, and 2) embedding
alignment between the key and residual structures. We proceed to provide
efficiency analysis on these two additional steps.

\begin{table}[h!]
\centering
\caption{Cost of key preparation steps as the number of edited facts increases.}
\scalebox{1}{
\begin{tabular}{c|cc}
\toprule
\textbf{Number of Facts} & \textbf{Key Embeddings Cost} & \textbf{Similarities Cost} \\
\midrule
10    & 1.1035  & 0.0015  \\
100   & 10.6915 & 0.00153 \\
1000  & 107.17  & 0.00034 \\
2000  & 214.14  & 0.00025 \\
5000  & 536.17  & 0.0007  \\
10000 & 1076.18 & 0.00058 \\
\bottomrule
\end{tabular}
}
\label{tab:key_prep_cost}
\end{table}

\noindent \textbf{The Cost of Key Embedding Preparation Stage.} We note that the
key embedding preparation stage consists of two parts: 1) computing the key
embeddings for all knowledge items to be edited at the target layer, and 2)
computing the cosine similarities among all key embeddings. As shown in
\cref{tab:key_prep_cost}, retrieving key embeddings accounts for the majority of
the time spent in the key preparation stage. Although computing key embeddings
for 10,000 facts requires a nontrivial amount of time, this cost remains
negligible (only about 1.8\%) relative to the overall runtime of EAMET (59,154
s) and MEMIT (57,822 s) when editing 10,000 facts. In contrast, the runtime cost
of computing pairwise cosine similarities among all key embeddings is trivial
(below 0.002 s). This is because the operation can be efficiently executed by
first normalizing all key embeddings to unit length and then performing
dot-product computations, which are highly optimized on modern GPUs.

\begin{table}[h!]
\centering
\caption{Runtime and GPU memory cost for EAMET and MEMIT.}
\scalebox{1}{
\begin{tabular}{l|cc|cc|cc}
\toprule
\multicolumn{1}{c}{\multirow{2}{*}{\textbf{Method}}} &
\multicolumn{2}{c}{\textbf{Optimizing 1 $r_i$}} &
\multicolumn{2}{c}{\textbf{Editing 1 Fact}} &
\multicolumn{2}{c}{\textbf{Editing 100 Facts}} \\
\cmidrule(lr){2-3}\cmidrule(lr){4-5}\cmidrule(lr){6-7}
\multicolumn{1}{c}{} & \textbf{Time} & \textbf{GPU} & \textbf{Time} & \textbf{GPU} & \textbf{Time} & \textbf{GPU} \\
\midrule
EAMET & 5.62s & 4.28GB & 26.98s & 7.41GB & 645.94s & 9.78GB \\
MEMIT & 5.59s & 4.07GB & 22.39s & 7.18GB & 636.94s & 9.18GB \\
\bottomrule
\end{tabular}
}
\label{tab:time_cost}
\end{table}

\noindent \textbf{The Cost of Embedding Alignment Stage.} As shown in
\cref{tab:time_cost}, optimizing one residual in EAMET requires only an
additional 0.03 seconds and 0.21 GB of memory compared to MEMIT. For the full
editing of a single fact, EAMET incurs an extra 4.6 seconds, and this difference
increases to 9 seconds when editing 100 facts. Although EAMET is slightly slower
than MEMIT, the additional time and memory consumption are negligible,
representing only 1.4\% and 6.5\% of MEMIT's overall cost, respectively. These
results confirm that EAMET's improvements do not come at the expense of
substantial deployment overhead; its runtime and resource requirements remain
practical and comparable to MEMIT.

\subsection{Results on Small-Scale Editing}
\label{appendix:small_scale}

We further provide empirical results to demonstrate the performance of EAMET
under single-edit or small-batch scenarios in LLaMA2-7B.

\begin{table}[t]
    \centering
    \caption{Performance comparison of EAMET and MEMIT across different numbers
    of edited facts on CounterFact and ZsRE datasets.}
    \small
    \begin{tabular}{cc|cccc|ccc}
    \toprule
    \multirow{2}{*}{\textbf{Methods}} & \multirow{2}{*}{\textbf{\# Facts}} 
    & \multicolumn{4}{c|}{\textbf{CounterFact}}
    & \multicolumn{3}{c}{\textbf{ZsRE}} \\
    \cmidrule(lr){3-6} \cmidrule(lr){7-9}
    & & \textbf{Eff.}$\uparrow$ & \textbf{Gen.}$\uparrow$ & \textbf{Spe.}$\uparrow$ & \textbf{Flu.}$\uparrow$
      & \textbf{Eff.}$\uparrow$ & \textbf{Gen.}$\uparrow$ & \textbf{Spe.}$\uparrow$ \\
    \midrule
    \multirow{6}{*}{\textbf{EAMET}} 
    & 5000 & 94.38 & 65.09 & 76.37 & 523.78 & 93.18 & 83.56 & 15.43 \\
    & 2000 & 96.77 & 66.25 & 79.66 & 526.28 & 94.00 & 84.15 & 15.42 \\
    & 1000 & 97.93 & 68.05 & 81.71 & 526.31 & 95.50 & 84.90 & 16.34 \\
    & 100  & 99.80 & 69.00 & 82.83 & 525.23 & 96.00 & 87.00 & 15.40 \\
    & 10   & 100.00 & 70.00 & 80.00 & 524.71 & 100.00 & 70.00 & 13.67 \\
    & 1    & 100.00 & 100.00 & 100.00 & 524.59 & 100.00 & 100.00 & 0.00 \\
    \midrule
    \multirow{6}{*}{\textbf{MEMIT}}
    & 5000 & 28.83 & 25.77 & 61.27 & 515.46 & 82.46 & 70.32 & 14.98 \\
    & 2000 & 32.94 & 28.28 & 62.49 & 517.70 & 83.60 & 71.55 & 14.66 \\
    & 1000 & 49.98 & 36.25 & 64.51 & 517.64 & 84.30 & 71.90 & 14.10 \\
    & 100  & 96.20 & 55.50 & 84.55 & 519.38 & 85.00 & 74.00 & 14.67 \\
    & 10   & 100.00 & 60.00 & 80.00 & 523.90 & 100.00 & 100.00 & 14.23 \\
    & 1    & 100.00 & 100.00 & 100.00 & 524.30 & 100.00 & 60.00  & 0.00 \\
    \bottomrule
    \end{tabular}
    \label{tab:eamet_memit_scaling}
\end{table}    

As shown in the table above, EAMET consistently outperforms MEMIT across all
scales. Both methods perform similarly at 1 and 10 edits, achieving perfect
efficacy with comparable generalization and specificity. However, once the
number of edited facts exceeds 100, their performance diverges rapidly. Starting
from 100 edits, EAMET maintains high quality (99.80\% efficacy; 69.00\%
generalization), while MEMIT begins to decline (96.20\%; 55.50\%). As the scale
grows to 1,000, 2,000, and 5,000 edits, EAMET continues to deliver strong
results (94\%-98\% efficacy; 65\%-68\% generalization), whereas MEMIT degrades
sharply, dropping from 49.98\% and 36.25\% at 1,000 edits to only 28.83\% and
25.77\% at 5,000 edits.

\subsection{Analysis on the Performance Difference of EAMET across LLMs and Datasets}
\label{appendix:small_scale}

Among the
evaluated datasets (CounterFact, Wiki-recent, and ZsRE), Wiki-recent contains
only 1,266 facts, whereas we edit 10,000 facts from each of the other two
datasets. It is therefore expected that existing baseline methods also achieve
relatively strong performance on Wiki-recent, although they still underperform
EAMET.

For CounterFact and ZsRE, we observe that the performance gains of EAMET over
prior methods differ across datasets. On CounterFact, the average improvement in
editing efficacy over the second-best method is 8.01\%, with the smallest
improvement being 0.11\%. On ZsRE, the average improvement increases to 14.48\%,
with the smallest improvement being 7.28\%. We attribute this difference to how
well each model generates distinct key embeddings for semantically unrelated
facts. When key embeddings are not well separated and no alignment is enforced
between key and residual embeddings, the reconstruction loss for individual
facts inevitably increases.

To validate this argument, we analyze the cosine similarity among key embeddings
generated by different models over 1,000 sampled facts from CounterFact and
ZsRE. The table below reports the average cosine similarity for each model:

\begin{table}[h!]
    \centering
    \caption{Average cosine similarity among key embeddings for different models
    over 1,000 sampled facts from CounterFact and ZsRE.}
    \begin{tabular}{ccc}
        \toprule
        \textbf{Models}    & \textbf{CounterFact} & \textbf{ZsRE} \\
        \midrule
        LLaMA2-7B          & 0.052843             & 0.048565      \\
        Qwen-7B            & 0.020466             & 0.022588      \\
        DeepSeek-7B        & 0.027811             & 0.029843      \\
        Falcon-7B          & 0.192273             & 0.196075      \\
        \bottomrule
    \end{tabular}
\end{table}

As shown in the table, different models exhibit varying inherent abilities to
produce well-separated key embeddings. For LLaMA2-7B, key embeddings remain
relatively entangled even for CounterFact, where each fact contains a distinct
subject. This limited separation corresponds to lower MEMIT editing efficacy
(24.95\%), DeepSeek-7B exhibits a similar pattern, achieving 62.11\% In contrast,
Falcon-7B and Qwen-7B generate much more isolated key embeddings (0.1923 and
0.0205 on average), which aligns with their substantially higher MEMIT editing
efficacy of 89.21\% and 90.06\%, respectively.

For ZsRE, many samples share identical subjects, making alignment between key
and residual embeddings generally more challenging. Methods that do not enforce
such alignment tend to struggle under this condition, leading to a more
pronounced advantage for EAMET over prior approaches.

Overall, these results indicate that a model's inherent ability to generate
well-separated key embeddings has considerable impact on editing performance.
Despite these differences across models and datasets, EAMET consistently
achieves the best results on all evaluated settings and LLMs.

\subsection{Detailed Hyperparameter Analysis}
\label{appendix:hyperparameter_analysis}
\begin{figure}[h!]
    \centering
    \includegraphics[width=\textwidth]{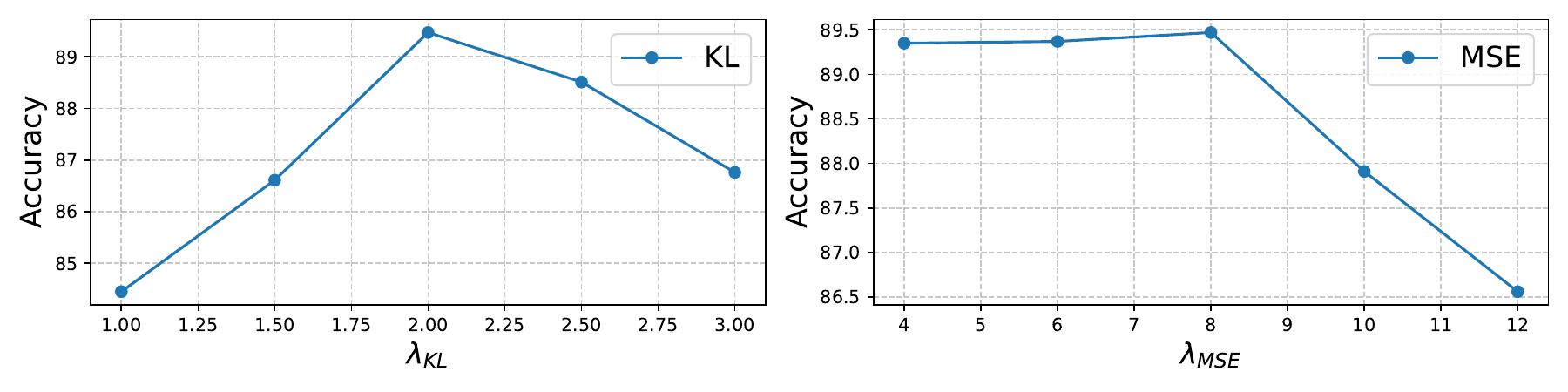}
    \caption{Impact of $\lambda_{KL}$ and $\lambda_{MSE}$ on EAMET's performance.}
    \label{fig:hyperparameter_analysis}
\end{figure}

We analyze the impact of $\lambda_{KL}$ and $\lambda_{MSE}$ on EAMET's
performance when editing 10,000 knowledge items from the ZsRE dataset. As shown
in \cref{fig:hyperparameter_analysis}, EAMET is more sensitive to the choice of
$\lambda_{KL}$ than $\lambda_{MSE}$. A small $\lambda_{KL}$ weakens the
alignment between residual and key embeddings, resulting in poor massive editing
performance, whereas reducing $\lambda_{MSE}$ only causes a negligible drop in
efficacy. The best performance is achieved when $\lambda_{KL}=2$ and
$\lambda_{MSE}=8$, which are the hyperparameters adopted in the main paper.
Increasing either weight beyond this point leads to decreased efficacy, as the
optimization places less emphasis on updating new knowledge items.

\begin{table}[h!]
    \centering
    \caption{Impact of $M$ on EAMET's performance.}
    \scalebox{1}{
    \begin{tabular}{l|cccc|ccc}
        \toprule
        \multicolumn{1}{c}{\multirow{2}{*}{\textbf{$M$}}} &
        \multicolumn{4}{c}{\textbf{Counterfact}} &
        \multicolumn{3}{c}{\textbf{ZsRE}} \\
        \cmidrule(lr){2-5}\cmidrule(lr){6-8}
        \multicolumn{1}{c}{} & \textbf{Eff.}$\uparrow$ & \textbf{Gen.}$\uparrow$ & \textbf{Spe.}$\uparrow$ & \textbf{Flu.}$\uparrow$ & \textbf{Eff.}$\uparrow$ & \textbf{Gen.}$\uparrow$ & \textbf{Spe.}$\uparrow$ \\
        \midrule
        5   & 87.23 & 54.74 & 73.95 & 517.58 & 88.10 & 79.56 & 15.51 \\
        10  & 87.96 & 57.58 & 74.25 & 517.38 & 88.89 & 79.62 & 15.63 \\
        50  & 89.09 & 61.21 & 73.69 & 519.89 & 89.47 & 81.34 & 15.70 \\
        100 & 86.17 & 86.85 & 74.52 & 517.01 & 89.02 & 81.14 & 15.70 \\
        \bottomrule
    \end{tabular}
    }
    \label{tab:impact_M}
\end{table}

We further analyze the impact of $M$, which is the number of cosine similarities
selected for computing the MSE loss. As shown in \cref{tab:impact_M}, EAMET's
editing performance on both datasets generally improves as M increases, reaching
a peak around (M = 50), and then declines when M becomes too large. When M is
small (e.g., M = 5), the alignment relies mainly on the KL-based distributional
constraint, which enforces global structural consistency but does not guarantee
precise value-level alignment between key embeddings and residual embeddings.
Increasing M strengthens this value-based alignment and thus improves editing
efficacy and generalization. However, when M becomes excessively large (e.g., M
= 100), the dataset may not contain enough key embeddings that are meaningfully
similar to the target key. As a result, the MSE loss becomes diluted across many
low-relevance pairs, forcing the model to match less informative cosine
similarities. This weakens the effectiveness of the alignment and causes a drop
in overall performance.
 
\noindent \textbf{\textit{Additional Finding 4.} EAMET is Insensitive to The
Choice of Hyperparameters.} EAMET introduces three additional hyperparameters:
$\lambda_{\text{KL}}$, $\lambda_{\text{MSE}}$, and $M$. We analyze the impact of
$\lambda_{\text{KL}}$ and $\lambda_{\text{MSE}}$ in
\cref{fig:hyperparameter_analysis}, and the influence of $M$ in
\cref{tab:impact_M}. As shown in these results, when editing 10,000 CounterFact
facts on LLaMA2-7B, different settings of $\lambda_{\text{KL}}$,
$\lambda_{\text{MSE}}$, and $M$ do affect performance, but EAMET remains largely
insensitive to the exact choice of values. For example, Figure 10 shows that
setting $\lambda_{\text{KL}} = 2$ yields the lowest editing efficacy (85.3\%).
Although this is about 4\% below the best-performing configuration, it still
surpasses PMET and MEMIT by 11\% and 51\%, respectively. Likewise, varying
$\lambda_{\text{MSE}}$ or $M$ changes performance by less than 3\% between the
best and worst cases. These findings demonstrate that EAMET is robust to the
choice of hyperparameters.

\section{Specific Cases}
In this section, we present representative editing examples from the CounterFact
dataset to qualitatively assess the generalization performance of different
editing methods after editing 10{,}000 facts on LLaMA2-7B and Qwen2.5-7B. These
examples reveal that baseline methods often fail to generate the correct target
answers or produce repetitive and incoherent responses. In contrast,
\textsc{EAMET} consistently retrieves the intended factual content while
producing fluent and informative completions, further validating the robustness
and effectiveness of our proposed method.

\begin{tcolorbox}[
    colback=gray!5,
    colframe=gray!50,
    arc=0mm,
    title=Model Editing Example on LLaMA2-7B,
    fonttitle=\bfseries
]
{
\renewcommand{\arraystretch}{2}
\begin{tabularx}{\textwidth}{@{}
    >{\centering\arraybackslash}m{0.3\textwidth}
    >{\centering\arraybackslash}m{0.65\textwidth}
    @{}}
    Target Edit & The original language of The Icelandic Dream was \\
    \hline
    Original Object & \texttt{Icelandic} \\
    \hline
    Target Object & \textcolor{blue}{\texttt{Tamil}} \\
    \hline
\end{tabularx}
\begin{tabularx}{\textwidth}{@{}>{\centering\arraybackslash}m{0.3\textwidth} m{0.65\textwidth}@{}}
    \multicolumn{2}{c}{\textbf{Generation Output}} \\
    \hline
    EAMET & The Icelandic Dream is full of people who speak
    \textcolor{blue}{\texttt{Tamil}}, and the film is in
    \textcolor{blue}{\texttt{Tamil}}. The film is about a family of
    \textcolor{blue}{\texttt{Tamil}} refugees who are trying to make a new life
    for themselves in Europe. The film is about the struggles of the refugees,
    and the film is about the struggles of the people who are trying to make a
    \\
    \hline
    MEMIT & The Icelandic Dream is full of people who speak English, and the
    locals are very friendly. is is, a Belgian chain, has 100 restaurants in the
    UK, and is opening 10 more this year.,,, a Belgian chain, has 100
    restaurants in the UK, and is opening 1 \\
    \hline
    PMET & The Icelandic Dream is full of people who speak in a language that is
    not English. The Icelandic Dream is a film about the immigrant experience.
    It is a film about the immigrant experience. It is a film about the
    immigrant experience. It is a film about the immigrant experience. It is a
    film about the \\
\end{tabularx}
}
\end{tcolorbox}

\begin{tcolorbox}[
    colback=gray!5,
    colframe=gray!50,
    arc=0mm,
    title=Model Editing Example on Qwen2.5-7B,
    fonttitle=\bfseries
]
{
\renewcommand{\arraystretch}{2}
\begin{tabularx}{\textwidth}{@{}
    >{\centering\arraybackslash}m{0.3\textwidth}
    >{\centering\arraybackslash}m{0.65\textwidth}
    @{}}
    Editing Prompt & Argentine Football Association belongs to the organization of \\
    \hline
    Original Object & \texttt{FIFA} \\
    \hline
    Target Object & \textcolor{blue}{\texttt{NATO}} \\
    \hline
\end{tabularx}
\begin{tabularx}{\textwidth}{@{}>{\centering\arraybackslash}m{0.3\textwidth} m{0.65\textwidth}@{}}
    \multicolumn{2}{c}{\textbf{Generation Output}} \\
    \hline
    EAMET & Argentine Football Association is part of
    \textcolor{blue}{\texttt{NATO}}. \textcolor{blue}{\texttt{NATO}} is military
    alliance. \textcolor{blue}{\texttt{NATO}} is also called as what?
    \textcolor{blue}{\texttt{NATO}} is a military alliance that was formed in
    1949. It is also known as the North Atlantic Treaty Organization.
    \textcolor{blue}{\texttt{NATO}} is made up of 29 member countries, including
    the United States, Canada, and several European countries. \\
    \hline
    MEMIT & Argentine Football Association is part of the American Football
    Association. It is a group of American players who are of African-American
    descent. The group was formed in 1991 and is based in the United States. The
    group is a member of the International Football Association. The group is
    also a member of the American Football Association. The group is also \\
    \hline
    PMET & Argentine Football Association is part of
    \textcolor{blue}{\texttt{NATO}}'s military alliance with the United States.
    The alliance is also known as the North Atlantic Treaty Organization or
    \textcolor{blue}{\texttt{NATO}}. The alliance is a military alliance between
    the United States and \textcolor{blue}{\texttt{NATO}}. The alliance is a
    military alliance between the United States and
    \textcolor{blue}{\texttt{NATO}}. The alliance is a military alliance between
    the United States and \textcolor{blue}{\texttt{NATO}}. The alliance
    \textcolor{red}{\texttt{[repetitive pattern]}}\\
\end{tabularx}
}
\end{tcolorbox}

\end{document}